\documentclass{SCIS2026}

\begin{document}
\ArticleType{RESEARCH PAPER}
\Year{2025}
\Month{January}
\Vol{68}
\No{1}
\DOI{}
\ArtNo{}
\ReceiveDate{}
\ReviseDate{}
\AcceptDate{}
\OnlineDate{}
\AuthorMark{}
\AuthorCitation{}

\title{Unconstrained Multi-view Human Pose Estimation with Algebraic Priors}{Title for citation}

\author[1,2]{Xiaolin Qin}{{qinxl2001@126.com}}
\author[1,2]{Qianlei Wang}{}
\author[1, 2]{Jiacen Liu}{}
\author[3]{Chaoning Zhang}{}
\author[4]{Fei Zhu}{}
\author[5]{Zhang Yi}{{zhangyi@scu.edu.cn}}


\address[1]{Chengdu Institute of Computer Applications, Chinese Academy of Sciences, Chengdu 610213, China}
\address[2]{School of Computer Science and Technology, University of Chinese Academy of Sciences, Beijing 101408, China}
\address[3]{School of Computer Science and Engineering, University of Electronic Science and Technology of China, Chengdu 610054, China.}
\address[4]{Centre for Artificial Intelligence and Robotics at Hong Kong Institute of Science $\mathbb{\&}$ Innovation, \\ Chinese Academy of Sciences, Hong Kong 999077, China.}
\address[5]{School of Computer Science, Sichuan University, Chengdu 610065, China}

\abstract{
Recovering 3D human pose from multi-view imagery typically relies on precise camera calibration, which is often unavailable in real-world scenarios, thereby severely limiting the applicability of existing methods. To overcome this challenge, we propose an unconstrained framework that synergizes deep neural networks, algebraic priors, and temporal dynamics for uncalibrated multi-view human pose estimation. First, we introduce the Triangulation with Transformer Regressor (TTR), which reformulates classical triangulation into a data-driven token fusion process to bypass the dependency on explicit camera parameters. Second, to explicitly embed the inherent algebraic relations of the multi-view variety into the learning process, we propose the Gr\"{o}bner basis Corrector (GC). This pioneering loss formulation enforces constraints derived from the multi-view variety to ensure the neural predictions strictly adhere to the laws of projective geometry. Finally, we devise the Temporal Equivariant Rectifier (TER), which exploits the equivariance property of human motion to impose temporal coherence and structural consistency, effectively mitigating scale ambiguity in uncalibrated settings. Extensive evaluations on standard benchmarks demonstrate that our framework establishes a new state-of-the-art for uncalibrated multi-view human pose estimation. Notably, our approach significantly closes the performance gap between calibration-free methods and fully calibrated oracles.}
\keywords{Human pose estimation, multi-view geometry, Gr\"{o}bner basis corrector, transformer regressor, temporal equivariant rectifier}

\maketitle

\section{Introduction}
Recovering 3D human spatial configurations from 2D imagery constitutes a fundamental challenge in computer vision. Solving this task is critical for a broad spectrum of applications, ranging from immersive virtual interactions and digital avatar animation to precise biomechanical diagnostics and sports analytics \cite{wang2021deep}. While deep learning has recently driven significant advancements in monocular 3D pose estimation, this task remains inherently ill-posed. Specifically, the inevitable loss of depth information during single-view projection induces severe spatial ambiguity, allowing multiple 3D kinematic states to mathematically map onto an identical 2D observation. Consequently, multi-view systems have emerged as the prevailing paradigm for high-fidelity reconstruction by aggregating complementary geometric cues from diverse perspectives to explicitly resolve these ambiguities.

The success of multi-view 3D human pose estimation relied heavily on the restrictive assumption of precise camera calibration. Most state-of-the-art frameworks require known intrinsic and extrinsic parameters. While this premise facilitates the direct application of epipolar geometry and volumetric triangulation, it creates a severe calibration bottleneck that significantly restricts applicability outside controlled laboratory environments. In real-world scenarios such as in-the-wild videos, archival footage analysis, and ad-hoc surveillance networks, calibration data is frequently unavailable, dynamic, or corrupted. Consequently, uncalibrated multi-view pose estimation necessitates the joint inference of the articulated 3D human structure and the complete camera geometry solely from 2D observations. This simultaneous recovery constitutes a highly complex instance of non-rigid Structure from Motion, which is inherently challenging to solve due to a highly non-convex optimization landscape and scale ambiguities that are further compounded by the complete absence of camera priors. Existing approaches to address these challenges generally fall into two distinct methodological paradigms, namely optimization-based geometry and data-driven learning, both of which exhibit fundamental limitations.

\begin{figure}[t!]
	\begin{center}
		\includegraphics[width = 1\linewidth]{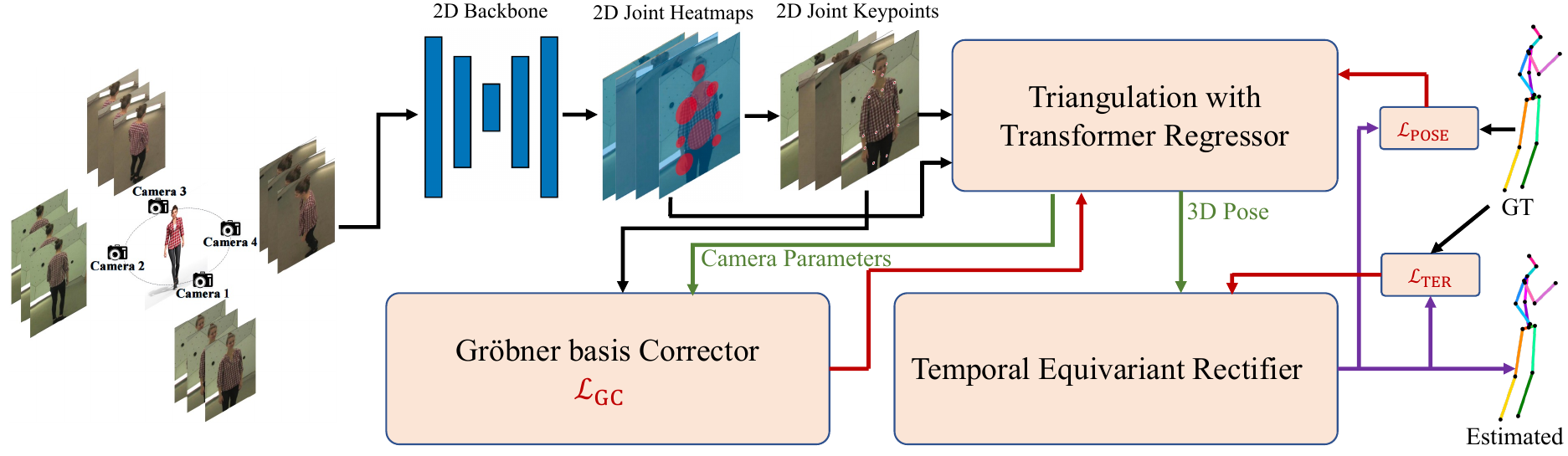}
	\end{center}
	\caption{Overview of our framework. It consists of three main modules: Triangulation with Transformer Regressor (TTR), Gr\"{o}bner basis Corrector (GC), Temporal Equivariant Rectifier (TER). The TTR is supervised using $\mathcal{L}_{\mathrm{GC}}$ and $\mathcal{L}_{\mathrm{POSE}}$. The Temporal Equivariant Rectifier is supervised using $\mathcal{L}_{\mathrm{TER}}$.}
	\label{all}
\end{figure}

Early multi-view approaches primarily relied on iterative Bundle Adjustment \cite{agarwal2010bundle, hartley2003multiple, geyer2001structure} and matrix factorization techniques that leverage rank constraints for articulated bodies \cite{bregler2000recovering, taycher2002recovering}. While mathematically elegant, these methods are computationally prohibitive for real-time applications and inherently fragile. Specifically, they demand clean 2D detections and exhibit high sensitivity to initialization, frequently falling into local minima that cause severe structural failures in completely uncalibrated settings \cite{akhter2008nonrigid}. To circumvent these bottlenecks, recent calibration-free deep learning paradigms \cite{xu2022multi, gordon2022flex, yin2025easyret3d, zhang2025efmk, li2024multi, song2026ectformer} bypass explicit geometric triangulation by employing neural networks for either latent feature aggregation \cite{usman2022metapose, jiang2023probabilistic} or canonical spatial alignment \cite{wandt2021canonpose, yu2022multiview, zhu2025muc}. Although these models demonstrate impressive robustness to noise and occlusion, their fundamental reliance on statistical correlation renders them geometric black boxes. Because they merely approximate causal geometric laws through data-driven fitting, they frequently generate visually plausible but mathematically invalid projections, such as inconsistent bone lengths or severe deviations from the underlying geometric manifold \cite{wandt2019repnet}. Lacking explicit algebraic constraints to disentangle the articulated human structure from latent multi-camera geometries, these pure regression networks fail to comprehend invariant projective relationships and merely memorize the spatial layouts of the training camera rigs \cite{rhodin2018learning, gordon2022flex}, which ultimately results in catastrophic generalization failures when deployed in novel, uncalibrated scenarios.

As shown in Figure \ref{all}, to overcome the limitations of uncalibrated setting, we propose an unconstrained framework guided by algebraic priors, comprising three novel components. First, the Triangulation with Transformer Regressor (TTR) circumvents the dependency on explicit projection matrices. By employing a token-based architecture, it leverages the global receptive field to implicitly learn the geometric mapping from 2D joint heatmaps to 3D space, functioning as a learnable triangulation engine robust to occlusion and calibration noise. Second, to ensure the estimated camera parameters strictly adhere to the laws of multi-view geometry, we introduce the Gr\"{o}bner basis Corrector (GC) to explicitly embed these algebraic priors into the optimization process. Advancing beyond the framework of the Euclidean reprojection error, the proposed GC minimizes the differentiable residual terms derived from the Gr\"{o}bner basis of the polynomial ideal encoding the fundamental matrix constraints, which inherently encapsulates the rank deficiency of the fundamental matrix. This formulation provides algebraic supervision that forces predictions onto the correct geometric manifold without ground-truth calibration. Finally, to mitigate scale ambiguity and instability in uncalibrated environments, the Temporal Equivariant Rectifier (TER) exploits the natural equivariance and structural continuity of human motion. In summary, our main contributions are:

\begin{itemize}
	\item We propose the Triangulation with Transformer Regressor (TTR), which reformulates classical triangulation as a data-driven token fusion process. This architecture circumvents the dependency on explicit camera parameters, enabling robust 3D reconstruction even under severe calibration noise.
	
	\item  We introduce the Gr\"{o}bner basis Corrector (GC), a pioneering loss formulation that bridges deep learning with Algebraic Geometry. By enforcing constraints derived from the ideal of the multi-view variety, GC ensures that neural predictions strictly adhere to the laws of projective geometry.
	
	\item We devise the Temporal Equivariant Rectifier (TER), a module that exploits the equivariance property of human motion. It effectively mitigates the ill-posed scale ambiguity in unconstrained settings by imposing temporal coherence and structural consistency across video sequences.
	
	\item Extensive evaluations on standard benchmarks demonstrate that our framework sets a new state-of-the-art for multi-view pose estimation, significantly closing the performance gap between uncalibrated methods and fully calibrated oracles.
\end{itemize}

\section{Related Works}
\subsection{Calibrated 3D Pose Estimation} 
Multi-view imagery resolves depth ambiguities inherent in single-view inputs, which makes high-precision 3D human pose estimation tractable in calibrated settings. Mainstream pipelines typically adopt a two-stage paradigm that extracts 2D keypoints using pretrained estimators \cite{chen2018cascaded, zheng2021weakly, ding2022hapnet} before performing cross-view 3D reconstruction. During fusion, point-based approaches \cite{qiu2019cross,bultmann2021real,dong2019fast} leverage explicit Euclidean coordinates to incorporate temporal cues and body priors seamlessly, whereas heatmap-based methods \cite{iskakov2019learnable} retain rich spatial evidence and uncertainty distributions through confidence maps at the cost of requiring more elaborate fusion mechanisms. Recent advancements have further strengthened cross-view reasoning within this framework. For instance, Liao \textit{et al.} \cite{liao2024multiple} embedded geometric constraints into cross-view Transformers to iteratively couple geometry and appearance for improved occlusion robustness, while Srivastav \textit{et al.} \cite{srivastav2024selfpose3d} utilized differentiable projection and heatmap rendering for closed-loop self-supervised training to eliminate ground-truth dependency. Other notable contributions include voxel-partitioned 3D spatial attention \cite{chen20243dsa}, explicit uncertainty modeling to improve robustness under out-of-distribution camera configurations \cite{davoodnia2024upose3d}, and Projective State Space (PSS) modules tailored for limited views \cite{chharia2025mv}. Despite these algorithmic innovations, their fundamental reliance on known and static camera parameters renders them highly sensitive to calibration errors, where minor inaccuracies inevitably amplify into systematic 3D deviations during geometric projection. This strict dependence imposes severe engineering burdens and necessitates strong synchronization assumptions, which severely restricts generalization under lens distortion, focal length variations, or sparse viewpoints. Consequently, even with sophisticated robustifications, these calibrated pipelines still suffer pronounced performance degradation in scenarios where camera parameters are unknown or dynamically changing.

\subsection{Uncalibrated 3D pose estimation} 
Uncalibrated 3D human pose estimation initially relied on explicit geometric optimization, such as formulating iterative solvers with reprojection errors and body priors \cite{takahashi2018human} or estimating cameras from independently inferred per-view poses \cite{lee2022extrinsic}. Other early efforts utilized time-invariant bone lengths for forward kinematics \cite{gordon2022flex} or candidate scoring networks \cite{bartol2022generalizable}, yet these approaches typically produce camera-dependent outputs or require post-hoc calibration to achieve global alignment. To strengthen cross-view coupling and reduce supervision, recent literature has shifted towards advanced neural fusions and self-supervised paradigms. Specifically, Jia \textit{et al.} \cite{jia2023delving} proposed an iterative Pixel-aligned feedback fusion to disentangle global orientation from vertex-level features, whereas Luvizon \textit{et al.} \cite{luvizon2022consensus} achieved uncalibrated fusion through consensus optimization that jointly estimates intrinsic and extrinsic parameters from inverse-projected depths. For enhancing generalization across unseen configurations, Cai \textit{et al.} \cite{cai2024poseirm} applied invariant risk minimization to learn camera-insensitive representations, and Zhang \textit{et al.} \cite{zhang2025esmformer} introduced relative attention paired with an error-aware self-supervised loss to adaptively filter reliable 2D evidence. Furthermore, Muller \textit{et al.} \cite{muller2025reconstructing} extended uncalibrated recovery to the scene level by leveraging anthropometric constraints to jointly reconstruct multi-person meshes, point clouds, and cameras within a unified metric space.

Despite these methodological advancements, uncalibrated multiview estimation remains identifiable only up to a similarity transform, which leaves absolute scale and world alignment inherently ambiguous and highly sensitive to keypoint noise. To mitigate this geometric instability, most models resort to temporal stabilization. However, this heavy reliance on temporal cues necessitates strict multi-view synchronization and long input sequences, which causes structural failures under fast motion, occlusions, or sparse viewpoints, while precluding single-frame applicability. Moreover, compensating for missing calibration through pseudo-labels, consensus fitting, or strong anthropometric priors frequently amplifies early estimation errors and introduces severe domain bias.

\section{Methodology}
This section details our proposed framework, which synergizes deep learning with algebraic geometry for uncalibrated multi-view human pose estimation. We begin by establishing the theoretical foundation in Subsection 3.1, introducing the algebraic representation of the multiview variety and its universal Gr\"{o}bner basis. Crucially, these mathematical results are not merely descriptive, they serve as the fundamental guiding principles for our network's optimization strategy. Following standard 2D feature extraction, we introduce the Triangulation with Transformer Regressor (TTR) in Subsection 3.3, which acts as a data-driven engine to implicitly infer initial 3D poses and global camera parameters. To ensure these neural predictions strictly obey the laws of projective geometry, Subsection 3.4 presents the Gr\"{o}bner basis Corrector (GC). The GC directly translates the theoretical theorems established in Section 3.1 into explicit, differentiable algebraic constraints that guide the TTR. Finally, Subsection 3.5 details the Temporal Equivariant Rectifier (TER), an independent module that leverages human motion kinematics to resolve scale ambiguities and enforce structural coherence across temporal sequences.
\subsection{Preliminary}
The key to human pose estimation is how to accurately locate the joints of the human body. We use two matrices $\bold{m} \in \mathbb{R}^{2 \times J} $ and $\bold{M} \in \mathbb{R}^{3 \times J}$ to represent 2D and 3D poses respectively, where the $J$ represents the number of joints. Specifically, we utilize $\bold{m}^{[i:j]}$ and $\bold{M}^{[i:j]}$ to represent the copy of the $i$-th row and $j$-th column of  $\bold{m}$ and $\bold{M}$, which serve as coordinates of 2D and 3D poses. According to camera model, the relationship between $\bold{m}^{[:j]}$ and $\bold{M}^{[:j]}$ can be described as $\bold{m}^{[:j]}= P\bold{M}^{[:j]}$, where $P \in  \mathbb{R}^{3 \times 4} $ is the camera matrix which is composed of intrinsic matrix $K$ and extrinsic matrix $[R|t]$. We assume that the camera matrix is unknown, then the pose estimation problem in multi-view condition can be described as:
\begin{equation}
\min_{P_{\mu},P_{\nu}, \mathbf{w}} E_{\mu, \nu \in \Omega} [f_{(P_{\mu},P_{\nu}|\mathbf{w})}(\bold m_{\mu}, \bold m_{\nu}) - \bold M],
\end{equation}
where the $\mu$ and the $\nu$ represent different two camera, and the $\Omega$ means camera set which contains more than 2 cameras. Additional, the $f_{(P_{\mu},P_{\nu}|\mathbf{w})}(\cdot)$ represent the estimation function under the self parameters $\mathbf{w}$ and camera matrix pair ($P_{\mu},P_{\nu}$). The $\bold m_{\mu}$ and $\bold m_{\nu}$ represent the projection of $\bold M$ to 2D pose through camera $\mu$ and $\nu$ respectively.

In order to facilitate the introduction of our work, we first re-describe the camera configuration in algebraic language. As is well known, each camera is represented by a $3 \times 4$ non-singular matrix which specifies a linear projection from $\mathbb{P}^3_{\mathbb{R}}$ to $\mathbb{P}^2_{\mathbb{R}}$. We define camera set $\Omega$ as multiview configuration, and the length of $\Omega$ is $n$ which is equal to the number of cameras. Specifically, the joint image space formed by the $n$ camera views constitutes the image of a multilinear rational mapping through $\Omega$:
\begin{equation}
\varphi_{\Omega}:  \mathbb{P}^3_{\mathbb{R}} \rightarrow (\mathbb{P}^2_{\mathbb{R}})^n, \mathbf{x} \mapsto\left(P_1 \mathbf{x}, P_2 \mathbf{x}, \ldots, P_n \mathbf{x}\right).
\end{equation} 
In \cite{aholt2013hilbert}, the Zariski closure of this image is called a multiview variety of $(P_1, P_2, \cdots, P_n)$,  which uses $\mathcal{V}_p$ to describe. \\
\textbf{Definition 3.1.} \textit{The camera configuration is generic if $4 \times 4$ minors of the $4 \times 3n$ matrix $\left[\begin{array}{llll}P_1^T & P_2^T & \cdots & P_n^T\end{array}\right]$ are non-zero.} 

Then, we model $\varphi_{\Omega}$ by follow equation:
\begin{equation}
 P_{(n,j)} \xi_{(n,j)} \\
 =\left[\begin{array}{cccccc}
P_1 & \widetilde{\bold m}^{[:j]}_1  & 0 & 0 & \cdots & 0 \\
P_2 & 0 & \widetilde{\bold m}^{[:j]}_2 & 0 & \cdots & 0 \\
P_3 & 0 & 0 & \widetilde{\bold m}^{[:j]}_3 & \cdots & 0 \\
\vdots & \vdots & \vdots & \vdots & \ddots & \vdots \\
P_n & 0 & 0 & 0 & \cdots & \widetilde{\bold m}^{[:j]}_n
\end{array}\right]
\left[\begin{array}{c}
\widetilde{\bold M}^{[:j]} \\
-\lambda_1 \\
-\lambda_2 \\
-\lambda_3 \\
\vdots \\
-\lambda_n
\end{array}\right] \\
 = 0,
\end{equation}
where $\widetilde{\bold m}^{[:j]}_{n} \in \varphi_{\Omega}(\widetilde{\bold M}^{[:j]})$, and $\lambda_n$ represents the perspective scale factor. Based on Formula (3), the $P_{(n)}$ has a non-trivial kernel since it contains the $\xi $, and hence the maximal minors of $P_{(n)}$ vanish on $\widetilde{\bold m}$. \\
\textbf{Definition 3.2.} \textit{For any subset $\sigma=\left\{\sigma_1, \ldots, \sigma_k\right\} \subseteq[n]$, when $k \ge 2$, the  matrix }
\begin{equation}
P_{(\sigma, j)} = \left[\begin{array}{ccccc}P_{\sigma_1} & \widetilde{\bold m}^{[:j]}_{\sigma_1}  & 0 & \ldots & 0 \\ P_{\sigma_2} & 0 & \widetilde{\bold m}^{[:j]}_{\sigma_2}& \ddots & 0 \\ \vdots & \vdots & \ddots & \ddots & \vdots \\ P_{\sigma_k} & 0 & \ldots & 0 & \widetilde{\bold m}^{[:j]}_{\sigma_k} \end{array}\right]
\end{equation}
\textit{is called the partially symbolic matrix of $P_{(n,j)}$.}

According to \cite{hartley2003multiple}, the maximal minors of $3k \times (4+k)$ matrix $P_{(\sigma)}$ is considered frequently to describe constraints between different views, and the constraints are multilinear. \\
\textbf{Definition 3.3.} \textit{For $k \ge 2$, the maximal $(4+k) \times (4+k)$ minor of $P_{(\sigma)}$, termed the k-focal polynomial of $P_{(\sigma)}$, is denoted by $\operatorname{minor}_k(P_{(\sigma)})$.}

Due to starting directly from $k$-focal polynomial, we will try to split it into some specific submatrices. The  $\operatorname{minor}_{k-1}(P_{(\sigma)})$ can be transfered to the $\operatorname{minor}_{k}(P_{(\sigma)})$ by adding one row from a extra camera $P_{\sigma _e}$ and the associated coordinate \cite{agarwal2019ideals}, which is shown as follow:
\begin{equation}
\operatorname{minor}_{k-1}(P_{(\sigma)})  \overset{\rhd (P_{\sigma_e}[i:],\widetilde{\bold m}^{[i:\sigma_e]})}{\mapsto}  \operatorname{minor}_{k}(P_{(\sigma)})
\end{equation}

If the $k$-focal polynomial obtained from the submatrices that include all the rows corresponding to two images and one row from each of the remaining images, then we will consider the $k$-focal polynomial satisfied bilinear constraints which more commonly known as epipolar constraints. The bilinear constraints can be represented as follow:
\begin{equation}
\mathbf{B}_{\mu,\nu}= \operatorname{det} \left[\begin{array}{lll}
P_{\mu} & \tilde{\mathbf{m}}_{\mu}^{[: j]} & 0 \\
P_{\nu} & 0 & \tilde{\mathbf{m}}_{\nu}^{[: j]}
\end{array}\right] = 0 .
\end{equation}

We use the linearity of the determinant to expand $	\mathbf{B}_{\mu,\nu}$ into terms with respect to $\tilde{\mathbf{m}}_{\mu}^{[: j]}$ and $\tilde{\mathbf{m}}_{\nu}^{[: j]}$:
\begin{equation}
\sum_{a=1}^3 \sum_{b=1}^3(-1)^{a+b} \operatorname{det}\left[\begin{array}{c}
P_{\mu}^{(a)} \\
P_{\nu}^{(b)}
\end{array}\right] \tilde{\mathbf{m}}_{\mu}^{[a : j]} \tilde{\mathbf{m}}_{\nu}^{[b : j]} 
\end{equation}
where the $P_{\mu}^{(a)}$ represents the submatrix after removing  $a$th column of $P_{\mu}$ and $P_{\nu}^{(b)}$ similarly. The $(-1)^{a+b} \operatorname{det}\left[\begin{array}{c}
P_{\mu}^{(a)} \\
P_{\nu}^{(b)}
\end{array}\right]$ forms the elements of the fundamental matrix $F_{\mu, \nu}$ which implicitly contains the relative motion and intrinsic information of the camera, then a more concise form can be obtained:
\begin{equation}
(\tilde{\mathbf{m}}_{\mu}^{[ : j]})^T F_{\mu, \nu} (\tilde{\mathbf{m}}_{\nu}^{[ : j]}) = 0
\end{equation}
\textbf{Definition 3.4.} \textit{A universal Gr\"{o}bner basis of an ideal $I$ is a subset that is a Gr\"{o}bner basis of the ideal under all term orders.} \\
\textbf{Theorem 3.5. }(Theorem 2.1 \cite{aholt2013hilbert}) \textit{If the camera configuration is generic, then the maximal minors of the matrices $P_{(\sigma)}$ for $2 \leq |\sigma| \leq 4$ form a universal Gr\"{o}bner basis of the multiview ideal $I_M$}.\\
\textbf{Proposition 3.6.} 
\textit{For a generic camera configuration (Definition 3.1) with camera set $\Omega$ of size $n$, let $\sigma \subseteq [n]$ be a view subset with $2\leq |\sigma|=k\leq4$, and $P_{(\sigma,j)}$ be the $3k\times(4+k)$ partially symbolic matrix for the $j$-th corresponding joint (Definition 3.2). The following statements hold:}
\begin{enumerate}
	\item \textit{The multi-view constraint system $P_{(\sigma,j)}\xi_{(\sigma,j)}=0$ admits a non-trivial solution if and only if all maximal $(4+k)$-order minors of $P_{(\sigma,j)}$ vanish. These vanishing minors collectively generate the multi-view ideal $I_M$, and for $2\leq k\leq4$, these minors exactly form a universal Gr\"{o}bner basis of $I_M$.}
	\item \textit{For any $\sigma$ with $2\leq k\leq4$, the maximal minor constructed via the row partition $(3,3,1,\dots,1)$ (full 3 rows from two distinct views in $\sigma$, 1 row from each of the remaining $k-2$ views) decomposes, up to a non-zero sign, into the product of a monomial in the image coordinates of the remaining $k-2$ views and a two-view bilinear constraint $(\tilde{m}_\mu^{[:j]})^T F_{\mu,\nu} \tilde{m}_\nu^{[:j]}=0$, which represents the core bilinear constraint of $I_M$.}
	\item \textit{For $k=3$ and $k=4$, maximal minors from balanced row partitions expand to multi-linear polynomials of multidegree $(1,1,1)$ and $(1,1,1,1)$ respectively. The coefficient vectors of these polynomials precisely span the row space of the Macaulay matrix constructed via the Segre embedding of the homogeneous image coordinates, thereby forming the high-order elements of the universal Gr\"{o}bner basis of $I_M$.}
\end{enumerate}

{\textit{Proof.}}

\begin{enumerate}
	\item {By Definition 3.2, the partially symbolic matrix $P_{(\sigma,j)}$ is of dimension $3k\times(4+k)$. The homogeneous linear system $P_{(\sigma,j)}\xi_{(\sigma,j)}=0$ admits a non-trivial kernel if and only if the $P_{(\sigma,j)}$ is rank-deficient. According to fundamental linear algebra, this is algebraically equivalent to the simultaneous vanishing of all its maximal $(4+k)$-order minors. Furthermore, following Theorem 3.5, for a generic camera configuration, the complete set of these maximal minors for $2\leq k\leq4$ forms the universal Gr\"{o}bner basis of the multi-view ideal $I_M$, fundamentally characterizing the multi-view variety $\mathcal{V}_p$.}
	\item {Given a view subset $\sigma$ with $|\sigma|=k\geq2$, the construction of a maximal minor via the highly skewed row partition $(3,3,1,\dots,1)$ involves the selection of exactly $k+4$ rows. Specifically, this selection incorporates all three rows from two designated views $\mu, \nu \in \sigma$ and a single row from each of the remaining $k-2$ views. It follows that the image coordinate columns corresponding to these $k-2$ single-row views contain a unique non-zero entry within the resulting submatrix. Applying Laplace expansion repeatedly along these specific columns extracts a sequence of non-zero monomial terms corresponding to the image coordinates of the respective views multiplied by a algebraic cofactor. Upon exhausting these $k-2$ columns, the remaining cofactor is a determinant formed exclusively by the full three rows of views $\mu$ and $\nu$. This determinant is precisely the bilinear epipolar constraint $\mathbf{B}_{\mu,\nu}$, which directly expands to the standard bilinear form $(\tilde{m}_\mu^{[:j]})^T F_{\mu,\nu} \tilde{m}_\nu^{[:j]}=0$.}
	\item {For the $k=3$ case, a balanced row partition such as $(3,2,2)$ selects a total of 7 rows. Because no view contributes only a single row, the selected submatrix intrinsically lacks the isolated single-entry columns requisite for the aforementioned monomial factorization. Expanding this determinant yields a homogeneous polynomial of multidegree $(1,1,1)$ with respect to the image coordinates of the three views, representing an trilinear constraint. By employing the standard Segre embedding of the homogeneous image coordinates, defined as $\mathfrak{m}_{\mu,\nu,\omega}(j) = \tilde{\mathbf{m}}_\mu^{[:j]} \otimes \tilde{\mathbf{m}}_\nu^{[:j]} \otimes \tilde{\mathbf{m}}_\omega^{[:j]}$, these trilinear polynomials act precisely as linear forms on the Segre embedding. The coefficient vectors of these polynomials precisely span the row space of the corresponding multi-view Macaulay matrix. A symmetric algebraic derivation directly applies to the $k=4$ case, where the balanced $(2,2,2,2)$ partition yields a minor that expands into an quadrilinear polynomial of multidegree $(1,1,1,1)$, corresponding to the row space of the Macaulay matrix acting on the Segre embedding. These multi-linear constraints serve as the essential high-order elements of the universal Gr\"{o}bner basis.} \hfill $\square$
\end{enumerate}

\subsection{Feature Extractor}
The feature extractor is divided into two parts. The first part is to obtain the heatmap through the pre-trained human keypoints detector backbone network. We followed the learnable triangulation \cite{iskakov2019learnable} to use the ``sample baseline" as our heatmap extractor:
\begin{equation}
\{ H_{\mu,1}, H_{\mu,2}, ..., H_{\mu,J} \} = Det_{2D}(I_{\mu}) \cdot Conv_{1 \times 1}
\end{equation}
where the $I_{\mu}$ represents the image of the $\mu$ camera, the $Det_{2D}$ represent ResNet-152 network with deconvolutional layers, through the $1 \times 1$ convolution we 
obtain heatmaps $\{ H_{\mu,1}, H_{\mu,2}, ..., H_{\mu,J} \}$ for all $J$ key points. The heatmaps will be used to generate the 3D pose coordinate $C_{3D}$ though our triangulation methods. At the same time, we obtain the confidence of each joint point $\{w_{\mu,1},w_{\mu,2},...,w_{\mu,J}\}$.

\subsection{Triangulation with Transformer Regressor}
\begin{figure}[t!]
	\begin{center}
		\includegraphics[width = 1\linewidth]{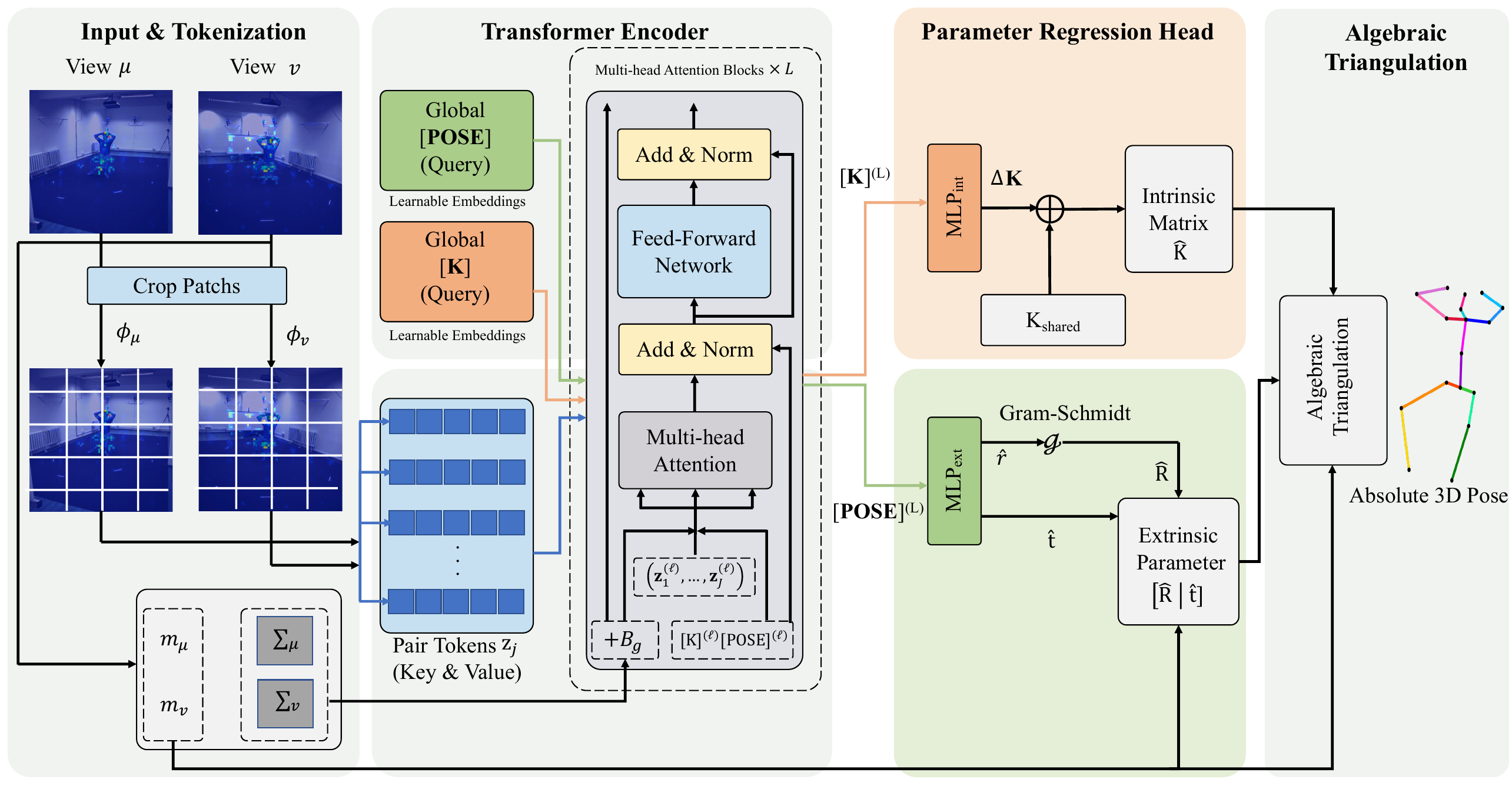}
	\end{center}
	\caption{Overall architecture of the Triangulation with Transformer Regressor (TTR). 
	}
	\label{TTR}
\end{figure}

The primary objective of the Triangulation with Transformer Regressor (TTR) is to bypass the traditional dependency on explicit camera calibration by formulating multi-view triangulation as a data-driven token fusion process. By implicitly learning the geometric mapping from 2D observations to 3D space, this module concurrently infers the initial 3D pose and the global camera parameters.
The internal architecture of the TTR is illustrated in Figure \ref{TTR}, where the input processing pipeline is defined as follows. Given a pair of views $(\mu, \nu)$ with matched joints and their heatmaps, we build a set of pair tokens from data and prepend two global tokens that are learnable parameters, denoted $\mathrm{[POSE]}$ and $\mathrm{[K]}$. The pair tokens encode visual evidence and epipolar geometry from the input and will serve as keys and values in the co-attention. Furthermore, the two global tokens function as learnable query embeddings to aggregate the extrinsic and intrinsic information of the view pair, respectively. Concretely, for each matched joint $j$, we crop $p \times p$ patches from the heatmaps around $\mathbf{m}_{\mu}^{[: j]}$ and $ \mathbf{m}_{\nu}^{[: j]}$ to obtain descriptors $\phi_{\mu, j}$ and $ \phi_{\nu, j}$. Then, we compose the $j$-th pair token as:
\begin{equation}
\begin{split}
\mathbf{z}_j = &  \big[  \phi_{\mu, j} \,\|\, \phi_{\nu, j} \,\|\, 
\tilde{\mathbf{m}}_{\mu}^{[0: j]}, \tilde{\mathbf{m}}_{\mu}^{[1: j]}, \tilde{\mathbf{m}}_{\nu}^{[0: j]}, \tilde{\mathbf{m}}_{\nu}^{[1: j]}, 
\tilde{\mathbf{m}}_{\mu}^{[0: j]} \tilde{\mathbf{m}}_{\nu}^{[0: j]}, \\ &\tilde{\mathbf{m}}_{\mu}^{[0: j]} \tilde{\mathbf{m}}_{\nu}^{[1: j]},
\tilde{\mathbf{m}}_{\mu}^{[1: j]} \tilde{\mathbf{m}}_{\nu}^{[0: j]}, 
\tilde{\mathbf{m}}_{\mu}^{[1: j]} \tilde{\mathbf{m}}_{\nu}^{[1: j]}, \log \operatorname{diag}\!\left(\Sigma_{\mu, j}+\Sigma_{\nu, j}\right)\big]
\end{split}
\end{equation}
We then form the input sequence:
\begin{equation}
\mathcal{Z}^{(\ell)}_g=\left(\mathrm{[POSE]}^{(\ell)},\mathrm{[K]}^{(\ell)}\right), \mathcal{Z}^{(\ell)}_p =\left( \mathbf{z}_1^{(\ell)}, \ldots, \mathbf{z}_J^{(\ell)} \right)
\end{equation}
A encoder with $L$ multi-head attention blocks processes $(\mathcal{Z}_g^{(0)}, \mathcal{Z}_p^{(0)}) \rightarrow (\mathcal{Z}_g^{(L)}, \mathcal{Z}_p^{(L)})$. To make the aggregation robust to outliers and heatmap blur, we add an uncertainty bias on keys that correspond to pair tokens:
\begin{equation}
\begin{split}
&\operatorname{Attn}_h\!\left(\mathcal{Z}_g^{(\ell)}, \mathcal{Z}_p^{(\ell)}\right) = \operatorname{softmax}\!\left(\frac{Q_h (K_h^{\top}+B_g)}{\sqrt{d_k}}\right) V_h, \\
&Q_h= \mathcal{Z}_g^{(\ell)} W_h^Q \in \mathbb{R}^{2 \times d_k}, \quad K_h= \mathcal{Z}_p^{(\ell)} W_h^K \in \mathbb{R}^{J \times d_k}, \quad V_h= \mathcal{Z}_p^{(\ell)} W_h^V \in \mathbb{R}^{J \times d_k}, \ell \in [0,1, \cdots, L],
\end{split}
\end{equation}
where $W_h^Q$, $W_h^K$ and $W_h^V$ represent learnable weights in $h$-th head, the $\Sigma_{\mu, j}$ and $\Sigma_{\nu, j}$ represent the 2D covariances, and the detail of bias $B_g \in \mathbb{R}^{2 \times J}$ is shown as follow:
\begin{equation}
\left(B_g\right)_{(\mu,\nu), j}=\log{\left(1 /\left(\operatorname{tr}\left(\Sigma_{\mu, j}+\Sigma_{\nu, j}\right)\right)\right)}, j \in [0,1,\cdots,J]
\end{equation}
The outputs from the $H$ attention heads are first concatenated. This combined representation is then added to the global token $\mathcal{Z}_g^{(\ell)}$ via a residual connection, followed by layer normalization to yield the intermediate feature $\mathcal{Z}_g^{(\ell)*}$.  To update the two global tokens, the $\mathcal{Z}_g^{(\ell)*}$ is processed by a feed-forward network and the second residual addition:

\begin{equation}
\left[ [\mathrm{POSE}]^{(\ell+1)} , [\mathrm{K}]^{(\ell+1)} \right] =  \mathrm{FFN} \left(  \mathcal{Z}_g^{(\ell)*}  \right) + \mathcal{Z}_g^{(\ell)*}, \quad
\mathcal{Z}_g^{(\ell)*}=\!\left( \|_{h=1}^H \operatorname{Attn}_h^{\mathrm{MAB}}\!
\left(\mathcal{Z}_g^{(\ell)}, \mathcal{Z}_p^{(\ell)}\right) \right) + \mathcal{Z}_g^{(\ell)}.
\end{equation}
Finally, we use $\mathrm{[POSE]}^{(L)}$ and $[\mathrm{K}]^{(L)}$ as the input of two MLPs to predict the internal and external parameters respectively. Importantly, to establish a unified 3D space in uncalibrated settings, the external parameters are predicted as absolute poses defined in a global coordinate system. Specifically, to balance the stability of global calibration with the flexibility required for uncalibrated settings, we adopt a Shared with Per-pair Update ($S+\Delta$) strategy for the intrinsic estimation. To strictly enforce physical geometric constraints, the MLPs do not regress full unstructured matrices. Instead, $\operatorname{MLP}_{\text{int}}$ predicts a 4D residual vector $\Delta K = [\Delta f_x, \Delta f_y, \Delta c_x, \Delta c_y]^{\top}$ to update only the valid focal length and principal point components of a learnable shared base matrix $K_{\text{shared}}$, strictly preserving the required upper-triangular structure. Similarly, to ensure the rotation matrix strictly lies on the $SO(3)$ manifold, $\operatorname{MLP}_{\text{ext}}$ predicts a 3D translation vector $\hat{t}$ and a 6D continuous rotation representation $\hat{r}$, which is mapped to a valid orthogonal matrix $\hat{R}$ via Gram-Schmidt orthogonalization. The estimation process can be formulated as:
\begin{equation}
\hat{K} = K_{\text{shared}} \oplus \operatorname{MLP}_{\text{int}}\left([\mathrm{K}]^{(L)}\right), \quad
[\hat{R} | \hat{t}] = \big[ \text{GS}(\hat{r}) \,|\, \hat{t} \big], \quad  [\hat{r} , \hat{t}] = \operatorname{MLP}_{\text{ext}}\left([\mathrm{POSE}]^{(L)}\right),
\end{equation}
where $\oplus$ denotes the parameter-specific addition to the intrinsic matrix, and $\text{GS}(\cdot)$ represents the Gram-Schmidt process. This formulation anchors the intrinsic estimation around a learnable canonical camera configuration ($K_{\text{shared}}$) to ensure optimization stability, while allowing dynamic instance-specific refinement ($\Delta K$) for arbitrary uncalibrated inputs. After obtaining the global camera parameter information, the algebraic method in \cite{iskakov2019learnable} is used to estimate the absolute 3D coordinates of the joint. 

To explicitly supervise the TTR module, we introduce a 3D pose reconstruction loss, $\mathcal{L}_{\text{POSE}}$, applied directly in the absolute global coordinate space. After the TTR module predicts the global camera parameters and performs differentiable algebraic triangulation, the resulting initial 3D skeleton is sequentially processed by the TER. This yields the final rectified 3D human pose $\widehat{\mathbf{M}} \in \mathbb{R}^{3 \times J}$. We define $\mathcal{L}_{\text{pose}}$ as the $L_1$ distance between the rectified joint coordinates $\widehat{\mathbf{M}}$ and the ground-truth 3D pose $\mathbf{M}^{*}$:
\begin{equation}
	\mathcal{L}_{\text{POSE}} = \frac{1}{J} \sum_{j=1}^{J} \left\| \widehat{\mathbf{M}}^{[:j]} - \mathbf{M}^{*[:j]} \right\|_1
\end{equation}
where $J$ denotes the total number of body joints, and $\mathbf{M}^{*[:j]}$ represents the 3D coordinate vector of the $j$-th ground-truth joint. In uncalibrated settings, neural networks are prone to outputting physically implausible camera configurations that overfit the 2D observations but violate the fundamental laws of multi-view projective geometry. To ensure that the predicted global intrinsic ($\hat{K}$) and extrinsic parameters ($\hat{R}, \hat{t}$) strictly adhere to these geometric constraints, we apply the $\mathcal{L}_{\text{GC}}$ which is presented in next section. Functioning as algebraic constraints, $\mathcal{L}_{\text{GC}}$ evaluate the polynomial residuals derived from the universal Gr\"{o}bner basis of the multi-view ideal. 

\subsection{Gr\"{o}bner basis Corrector}
While the TTR effectively regresses the initial spatial configurations, pure neural networks are prone to generating physically implausible camera parameters in the absence of ground-truth calibration. To address this, we introduce the Gr\"{o}bner basis Corrector (GC) to serve as a rigorous geometric supervisor, translating theoretical algebraic priors into explicit constraints that force the predicted cameras to inherently obey the laws of multi-view projective geometry. Based on Formula (8), Proposition 3.6, and Theorem 3.5, we construct the GC to explicitly enforce algebraic priors, thereby implicitly removing the strict constraints on known camera parameters. The starting point for these algebraic priors is the bilinear constraint, whose coefficients assemble the fundamental matrix $F_{\mu, \nu}$ and yield $(\tilde{\mathbf{m}}_{\mu}^{[ : j]})^T F_{\mu, \nu} (\tilde{\mathbf{m}}_{\nu}^{[ : j]}) = 0$. Proposition 3.6 states that, for any camera subset $\sigma$ with $|\sigma|=k \geq 2$, the vanishing of all $(4+k)$-minors of the partially symbolic block matrix $P_{(\sigma, j)}$ evaluated at the observed homogeneous image points $\left\{\tilde{\mathbf{m}}_{\sigma}^{[ : j]}\right\}$ is equivalent to the existence of a single 3D point and per-view scales that satisfy the stacked system $P_{(\sigma, j)} \xi_{(\sigma, j)}=0$. Theorem 3.5  further proves that, for a generic camera configuration, the complete set of these maximal minors for $2\leq k\leq4$ forms the universal Gr\"{o}bner basis of the multiview ideal $I_M$. Building on these theoretical results, we construct a set of differentiable losses that simultaneously project the regressed cameras onto the underlying multiview variety and extend the algebraic consistency from two-view geometry into the three-view and four-view settings.

For two views $\sigma=\{\mu, \nu\}$, let $(\hat{K}, \hat{R}, \hat{t})$ be the camera parameters regressed by the TTR, and let $\hat{F}_{\mu,\nu}=\hat{K}^{-\top}[\hat{t}_{\mu,\nu}]_{\times} \hat{R}_{\mu,\nu} \hat{K}^{-1}$ be the fundamental matrix. Given matched homogeneous pixel coordinates $(\tilde{\mathbf{m}}_{\mu}^{[ : j]}, \tilde{\mathbf{m}}_{\nu}^{[ : j]})$, define the design matrix $\tilde{A} \in \mathbb{R}^{N \times 9}$ row-wise by $(\tilde{A})_{i:}= \mathbf{a}_i^{\top}$, where $\mathbf{a}_i$ is the bilinear row from Formula (8). Rather than relying on unstable parametric elimination, such as extracting non-unique singular vectors of $\tilde{A}$ to solve a univariate eliminant, we construct the two-view Gr\"{o}bner basis residual directly within the ambient polynomial ring. By defining the ideal as $\langle \{ \mathbf{a}_i^\top \operatorname{vec}(F) \}_{i=1}^N, \operatorname{det}(F) \rangle$, we leverage its Gr\"{o}bner basis, which inherently encapsulates both the bilinear forms and the cubic determinantal polynomial. We define the two-view Gr\"{o}bner basis loss by the degree-normalized penalty:
\begin{equation}
\mathcal{L}_{\mathrm{GB}-2} = \frac{\|\tilde{A} \operatorname{vec}(\hat{F}_{\mu,\nu})\|_2^2}{\|\hat{F}_{\mu,\nu}\|_F^2} + \frac{|\det(\hat{F}_{\mu,\nu})|^2}{\|\hat{F}_{\mu,\nu}\|_F^6} + \varepsilon, 
\end{equation}
where $\|\cdot\|_F$ denotes the Frobenius norm to ensure scale invariance, and $\varepsilon$ is a strictly positive infinitesimal constant used to maintain the stability of the training process. While the explicit parameterization of $\hat{F}_{\mu,\nu}$ structurally guarantees that $\det(\hat{F}_{\mu,\nu}) \equiv 0$ due to the strict skew-symmetry of $[\hat{t}_{\mu,\nu}]_{\times}$, we deliberately retain this determinantal penalty in the formulation for theoretical completeness to rigorously reflect the full generating set of the underlying multi-view ideal.

\begin{figure}[t!]
	\begin{center}
		\includegraphics[width = 1\linewidth]{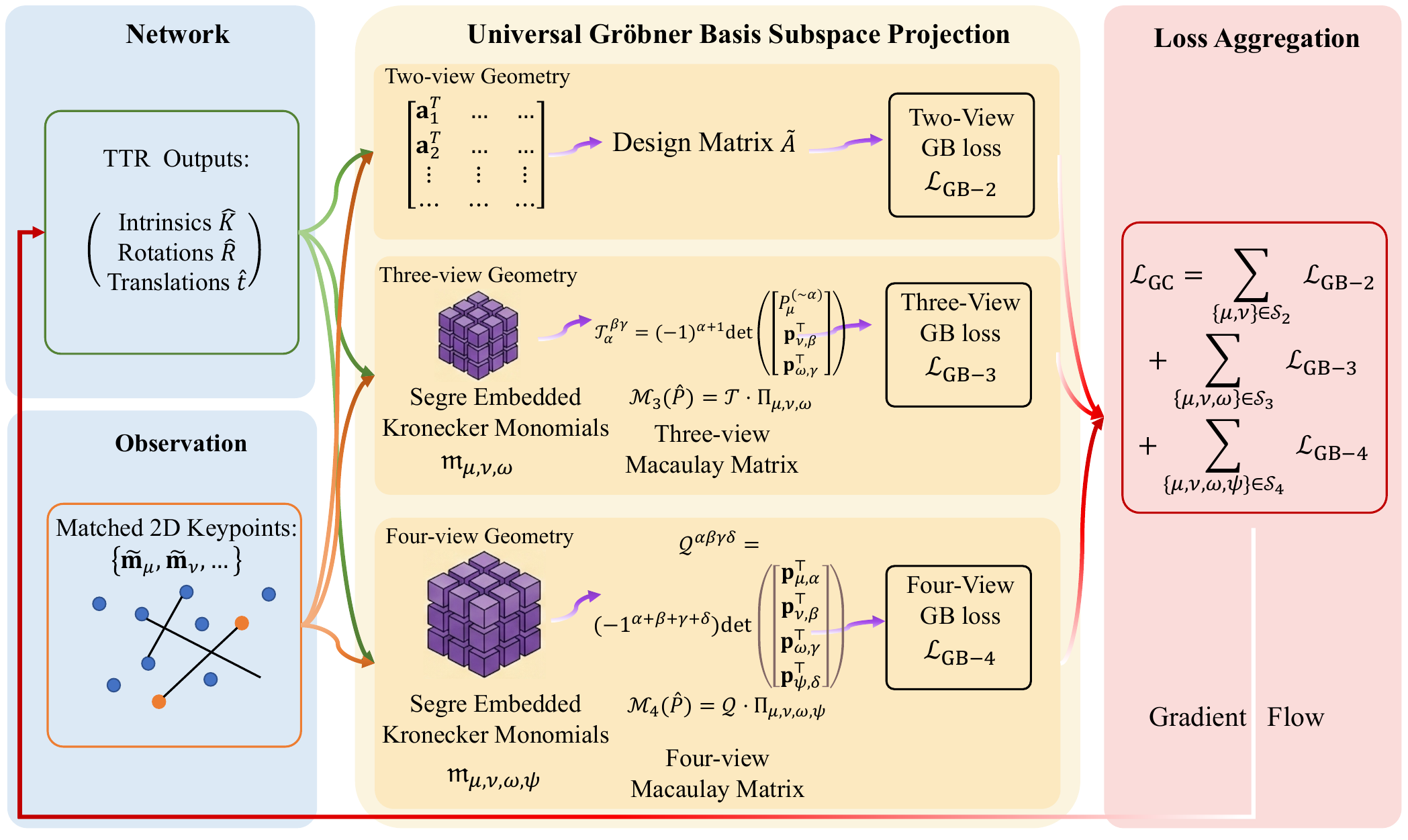}
	\end{center}
	\caption{Detailed architecture and gradient flow of the Gr\"{o}bner basis Corrector (GC).}
	\label{GC}
\end{figure}

For three views $\sigma=\{\mu, \nu, \omega\}$. By Proposition 3.6, the stacked system $P_{(\sigma,j)} \xi_{(\sigma,j)}=0$ has a nontrivial kernel if and only if all maximal $7$-order minors of the $9 \times 7$ partially symbolic matrix $P_{(\sigma,j)}$ vanish. Among these minors, the irreducible ones from the $(3,2,2)$ row partition expand to homogeneous trilinear polynomials of multidegree $(1,1,1)$, which are the trilinear minimal generators of the multiview ideal $I_M$. Concretely, define the Segre embedding coordinate vector:
\begin{equation}
\mathfrak{m}_{\mu, \nu, \omega}(j)=\tilde{\mathbf{m}}_{\mu}^{[: j]} \otimes \tilde{\mathbf{m}}_{\nu}^{[: j]} \otimes \tilde{\mathbf{m}}_{\omega}^{[: j]} \in \mathbb{R}^{27}.
\end{equation}

We first define the trifocal tensor $\mathcal{T}$ associated with the predicted camera set $\hat{P}=\{P_\mu, P_\nu, P_\omega\}$. Its elements are formulated via the $4 \times 4$ maximal minors of the joint projection matrix, which is strictly equivalent to the coefficient expansion of the irreducible $7$-order maximal minors of $P_{(\sigma,j)}$. Let $P_{\mu}^{(\sim \alpha)} \in \mathbb{R}^{2 \times 4}$ denote the submatrix obtained by removing the $\alpha$-th row of $P_\mu$, and let $\mathbf{p}_{\nu, \beta}^\top$ and $\mathbf{p}_{\omega, \gamma}^\top \in \mathbb{R}^{1 \times 4}$ denote the $\beta$-th and $\gamma$-th rows of $P_\nu$ and $P_\omega$, respectively. The tensor elements are defined as:
\begin{equation}
\mathcal{T}_{\alpha}^{\beta \gamma} = (-1)^{\alpha+1} \det\left( \begin{bmatrix} P_{\mu}^{(\sim \alpha)} \\ \mathbf{p}_{\nu, \beta}^\top \\ \mathbf{p}_{\omega, \gamma}^\top \end{bmatrix} \right), \quad \alpha, \beta, \gamma \in \{1,2,3\}.
\end{equation}

This tensor encodes the joint geometric constraints across all three views, and its entries are exactly the coefficients of the monomials in the irreducible trilinear polynomials. At multidegree $(1,1,1)$, we form a Macaulay matrix whose row space encodes these irreducible trilinear constraints: 
\begin{equation}
\mathcal{M}_3(\hat{P}) = \boldsymbol{\mathcal{T}} \cdot \Pi_{\mu, \nu, \omega} \in \mathbb{R}^{9 \times 27},
\end{equation}
where $\boldsymbol{\mathcal{T}} \in \mathbb{R}^{9 \times 27}$ is the coefficient matrix of the complete set of trilinear polynomials derived from the trifocal tensor. Specifically, the rows of $\boldsymbol{\mathcal{T}}$ are systematically constructed by flattening the tensor contractions corresponding to the standard point-point-point incidence relations. $\Pi_{\mu, \nu, \omega}$ is a block-wise permutation matrix that aligns the coefficient order of the trilinear polynomials with the monomial order of the Segre embedding $\mathfrak{m}_{\mu, \nu, \omega}(j)$, ensuring each coefficient acts on the corresponding triad of image coordinates. For generic camera configurations, the row space of $\boldsymbol{\mathcal{T}}$ has rank 4, which exactly spans all linearly independent irreducible trilinear minimal generators of $I_M$. Each row corresponds to a trilinear constraint of multidegree $(1,1,1)$, with entries ordered to match the monomial structure of the trifocal constraint. 

The three-view residual:
\begin{equation}
\mathcal{R}_3(j)=\mathcal{M}_3(\hat{P}) \mathfrak{m}_{\mu, \nu, \omega}(j) \in \mathbb{R}^{9},
\end{equation}
stacks the values of the complete set of trilinear trifocal constraints. Under generic camera configurations, the row space of these constraints precisely spans the irreducible trilinear minimal generators of the universal Gr\"{o}bner basis of $I_M$ for $|\sigma|=3$, eliminating the redundancy of trivial lifting from pairwise bilinear constraints and avoiding the degree mismatch of naive determinantal minor evaluation. We therefore directly define the three-view Gr\"{o}bner basis loss:
\begin{equation}
\mathcal{L}_{\mathrm{GB-3}}=\sum_{j=1}^{J} \left(\left\|\mathcal{R}_3(j)\right\|_2^2 / \eta_{\mathrm{tri}}^2\right),
\end{equation}
where $\eta_{\mathrm{tri}}$ is the critical normalization factor designed to prevent trivial minimization caused by the homogeneous scale freedom of the camera matrices and the resulting trifocal tensor. Specifically, we compute $\eta_{\mathrm{tri}}=\|\mathcal{M}_3(\hat{P})\|_F$ as the Frobenius norm of the Macaulay matrix to standardize the linear subspace constraints. By dividing the residual by the norm of its coefficients, the objective function inherently achieves scale invariance. This elegantly approximates the geometric distance from the algebraic error, ensuring that the network optimization is driven by projective joint geometric consistency rather than minimizing the numerical magnitude of the regressed matrices.

For four views $\sigma=\{\mu, \nu, \omega, \psi\}$. In parallel, Proposition 3.6 dictates that the stacked system $P_{(\sigma,j)} \xi_{(\sigma,j)}=0$ has a nontrivial kernel if and only if all maximal $8$-order minors of the $12 \times 8$ partially symbolic matrix $P_{(\sigma,j)}$ vanish. Among these minors, the irreducible ones from the $(2,2,2,2)$ row partition expand to homogeneous quadrilinear polynomials of multidegree $(1,1,1,1)$. Concretely, define the Segre embedding coordinate vector:
\begin{equation}
\mathfrak{m}_{\mu, \nu, \omega, \psi}(j)=\tilde{\mathbf{m}}_{\mu}^{[: j]} \otimes \tilde{\mathbf{m}}_{\nu}^{[: j]} \otimes \tilde{\mathbf{m}}_{\omega}^{[: j]} \otimes \tilde{\mathbf{m}}_{\psi}^{[: j]} \in \mathbb{R}^{81}.
\end{equation}

We first define the quadrifocal tensor $\mathcal{Q}$ associated with the predicted camera set $\hat{P}=\{P_\mu, P_\nu, P_\omega, P_\psi\}$. Its elements are formulated via the $4 \times 4$ maximal minors of the joint projection matrix, which is strictly equivalent to the coefficient expansion of the irreducible $8$-order maximal minors of $P_{(\sigma,j)}$. The 81 tensor elements are defined as:
\begin{equation}
\mathcal{Q}^{\alpha\beta\gamma\delta} = (-1)^{\alpha+\beta+\gamma+\delta} \det\left( \begin{bmatrix} \mathbf{p}_{\mu, \alpha}^\top \\ \mathbf{p}_{\nu, \beta}^\top \\ \mathbf{p}_{\omega, \gamma}^\top \\ \mathbf{p}_{\psi, \delta}^\top \end{bmatrix} \right), \quad \alpha,\beta,\gamma,\delta \in \{1,2,3\}.
\end{equation}

This tensor encodes the ultimate joint geometric constraints across all four views, and its entries are exactly the coefficients of the monomials in  quadrilinear polynomials. At multidegree $(1,1,1,1)$, we form a Macaulay matrix whose row space encodes these quadrilinear constraints: 
\begin{equation}
\mathcal{M}_4(\hat{P}) = \boldsymbol{\mathcal{Q}} \cdot \Pi_{\mu, \nu, \omega, \psi} \in \mathbb{R}^{81 \times 81},
\end{equation}
where $\boldsymbol{\mathcal{Q}} \in \mathbb{R}^{81 \times 81}$ is the coefficient matrix of the complete set of quadrilinear polynomials derived from the quadrifocal tensor. Specifically, the rows of $\boldsymbol{\mathcal{Q}}$ are systematically constructed by flattening the tensor contractions corresponding to the standard point-point-point-point incidence relations. $\Pi_{\mu, \nu, \omega, \psi}$ is a block-wise permutation matrix that aligns the coefficient order of the quadrilinear polynomials with the monomial order of the Segre embedding $\mathfrak{m}_{\mu, \nu, \omega, \psi}(j)$.

The four-view residual:
\begin{equation}
\mathcal{R}_4(j)=\mathcal{M}_4(\hat{P}) \mathfrak{m}_{\mu, \nu, \omega, \psi}(j) \in \mathbb{R}^{81},
\end{equation}
stacks the values of the complete set of quadrilinear constraints. Under generic camera configurations, the row space of these constraints precisely spans the quadrilinear generators of the universal Gr\"{o}bner basis of $I_M$ for $|\sigma|=4$. We therefore directly define the four-view Gr\"{o}bner basis loss:
\begin{equation}
\mathcal{L}_{\mathrm{GB-4}}=\sum_{j=1}^J \left(\left\|\mathcal{R}_4(j)\right\|_2^2 / \eta_{\mathrm {quad}}^2\right),
\end{equation}
where $\eta_{\mathrm{quad}} = \|\mathcal{M}_4(\hat{P})\|_F$ is the normalization coefficient comparable to $\eta_{\mathrm{tri}}$, ensuring that the objective function inherently achieves scale invariance.

Finally, let $\mathcal{S}_2, \mathcal{S}_3, \mathcal{S}_4$ be sets of pairs, triples and quadruples sampled from $\Omega$. The Gr\"{o}bner basis Corrector aggregates:
\begin{equation}
\mathcal{L}_{\mathrm{GC}} = \sum_{(\mu,\nu) \in \mathcal{S}_2} \mathcal{L}_{\mathrm{GB}-2} + \sum_{\{\mu,\nu,\omega\} \in \mathcal{S}_3} \mathcal{L}_{\mathrm{GB}-3} + \sum_{\{\mu, \nu, \omega, \psi\} \in \mathcal{S}_4} \mathcal{L}_{\mathrm{GB}-4}.
\end{equation}

All components are differentiable end-to-end. Firstly, the computations of determinants and Kronecker product are inherently smooth operations. Moreover, the elements of $\mathcal{M}_3$ and $\mathcal{M}_4$ are multilinear polynomial functions of the camera projection matrices $\hat{P}$, hence depending smoothly and differentiably on the network outputs $(\hat{K}, \hat{R}, \hat{t})$. Furthermore, the residuals are polynomial in the homogeneous image coordinates $\tilde{\mathbf{m}}$. Accordingly, gradients of $\mathcal{L}_{\mathrm{GC}}$ propagate smoothly through the Macaulay maps to the Transformer heads, where the [POSE] and [K] tokens are projected to $(\hat{R}, \hat{t})$ and $\hat{K}$, and further back to the attention mechanism over pair tokens. Ultimately, the loss $\mathcal{L}_{\mathrm{GC}}$ penalizes geometric violations, encouraging the regressed cameras to lie on the multiview variety defined by the universal Gr\"{o}bne basis. The complete pipeline is shown in Figure \ref{GC}.

The GC removes the need for ground-truth camera parameters or hard calibration ties during training, since supervision is supplied by polynomial vanishing sets rather than parameter equalities.  Thus only matched 2D evidence and heatmaps are required. The GC avoids hand-crafted priors on $(\hat{K},\hat{R},\hat{t})$ by steering them toward algebraic consistency, which tolerates moderate miscalibration and reduces bias from rigid calibration assumptions. It scales uniformly from pairs to triples and quadruples without redesigning solvers or changing parameterization, so the same objective adapts across rigs and view counts.

\subsection{Temporal Equivariant Rectifier}
\begin{figure}[t]
	\begin{center}
		\includegraphics[width = 1\linewidth]{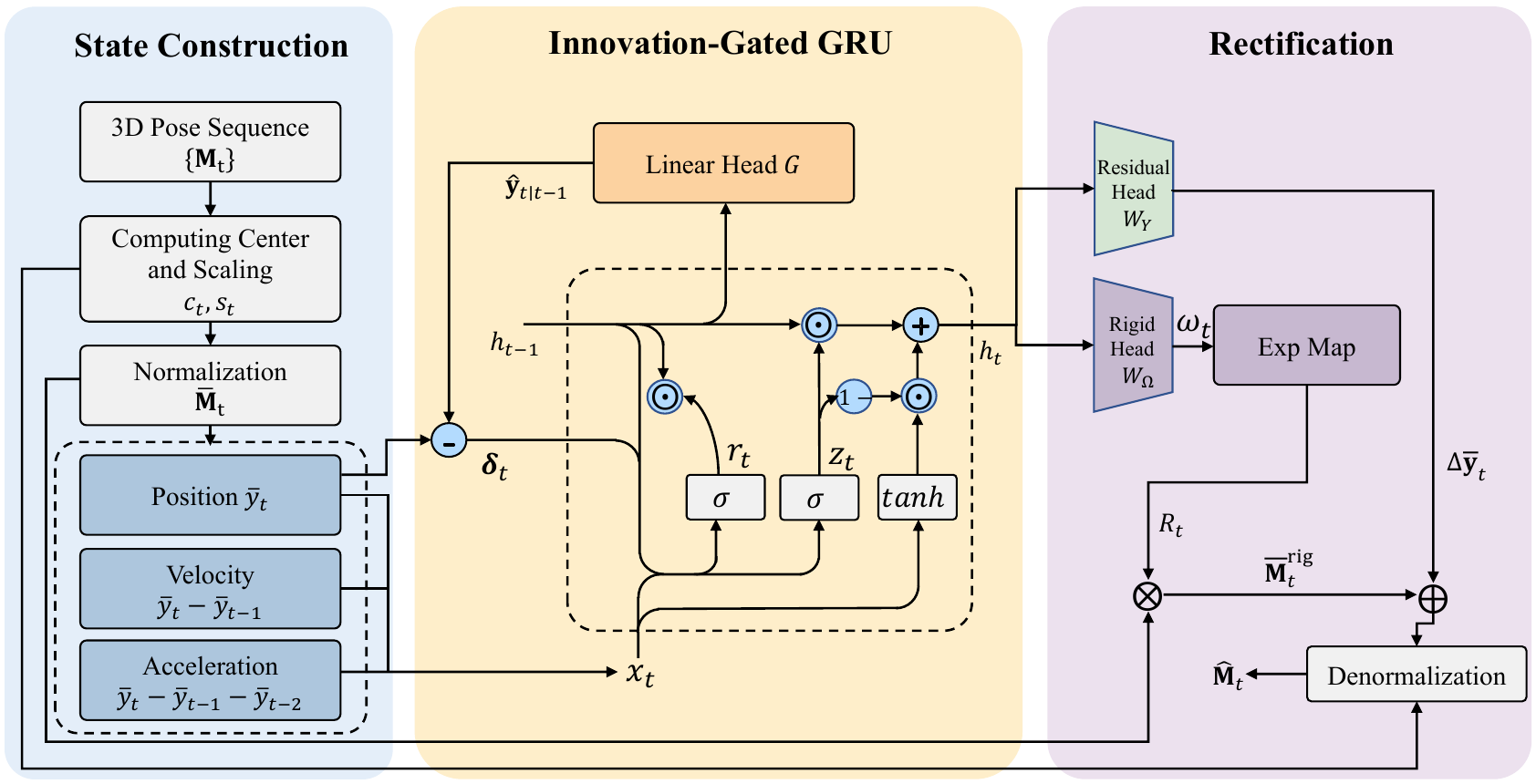}
	\end{center}
	\caption{Internal architecture and kinematic disentanglement workflow of the Temporal Equivariant Rectifier (TER). 
}
	\label{TER}
\end{figure}

To further refine the per-frame 3D poses generated by the TTR, we introduce a plug-and-play temporal corrector that is depicted in Figure \ref{TER}. The module is independent of the GC or camera constraints, and requires no kinematic or bone-length priors. Let $\mathbf{M}_t \in \mathbb{R}^{3 \times J}$ be the triangulated 3D pose at time $t$. To disentangle the articulated human motion from global trajectories, we first remove the global translation by subtracting the joint center $\mathbf{c}_t \in \mathbb{R}^{3}$ and normalize the per-frame scale:
\begin{equation}
\mathbf{c}_t=\frac{1}{J} \sum_{j=1}^J \mathbf{M}_{t}^{[:j]}, \quad \mathbf{\bar{M}}_t=\frac{\mathbf{M}_t-\mathbf{c}_t \mathbf{1}^{\top}}{s_t}
\end{equation}
where $s_t$ represents the mean of the Euclidean distances between all pairs of joints, and $\mathbf{\bar{M}}_t$ represents the normalized coordinate matrix. We then explicitly vectorize $\mathbf{\bar{M}}_t$ into a 1D state vector $\overline{\mathbf{y}}_t = \operatorname{vec}(\mathbf{\bar{M}}_t) \in \mathbb{R}^{3J}$ to serve as the sequential input.

The recurrent input aggregates position, articulated velocity, and articulated acceleration so that the model observes both the instantaneous pose configuration and short-term deformation dynamics:
\begin{equation}
\mathbf{x}_t=\left[\overline{\mathbf{y}}_t ; \overline{\mathbf{y}}_t-\overline{\mathbf{y}}_{t-1} ; \overline{\mathbf{y}}_t-2 \overline{\mathbf{y}}_{t-1}+\overline{\mathbf{y}}_{t-2}\right] \in \mathbb{R}^{9J}.
\end{equation}
For the initial frames ($t=1,2$), we replicate the first available observation to pad the missing history, and the initial hidden state $\mathbf{h}_0$ is initialized to zero. 

A gated recurrent unit maintains a hidden state $\mathbf{h}_t \in \mathbb{R}^{d_h}$. Furthermore, a linear head $G \in \mathbb{R}^{3J \times d_h}$ produces a one-step prior $\widehat{\mathbf{y}}_{t \mid t-1}=G \mathbf{h}_{t-1}$. The difference between the normalized observation and this prior, $\boldsymbol{\delta}_t=\overline{\mathbf{y}}_t-\widehat{\mathbf{y}}_{t \mid t-1}$, acts as an innovation signal that directly modulates the GRU gates:
\begin{equation}
z_t=\sigma\left(W_z \mathbf{x}_t+U_z \mathbf{h}_{t-1}+Q_z \boldsymbol{\delta}_t+b_z\right),
\end{equation}
\begin{equation}
r_t=\sigma\left(W_r \mathbf{x}_t+U_r \mathbf{h}_{t-1}+Q_r \boldsymbol{\delta}_t+b_r\right),
\end{equation}
\begin{equation}
\tilde{\mathbf{h}}_t=\tanh \left(W_h \mathbf{x}_t+U_h\left(r_t \odot \mathbf{h}_{t-1}\right)+b_h\right),
\end{equation}
\begin{equation}
\mathbf{h}_t=\left(1-z_t\right) \odot \mathbf{h}_{t-1}+z_t \odot \tilde{\mathbf{h}}_t,
\end{equation}
with learnable parameters $W_{\{\cdot\}}, U_{\{\cdot\}}, b_{\{\cdot\}}$, gate projection matrices $Q_{\{\cdot\}} \in \mathbb{R}^{d_h \times 3J}$, and element-wise product $\odot$. The innovation signal operates akin to a learned Kalman filter, allowing the model to automatically down-weight noisy frames while preserving reliable dynamics when observations are stable.

From $\mathbf{h}_t$, we predict a small rigid motion and a non-rigid residual. The rigid part is parameterized strictly by a rotation vector $\boldsymbol{\omega}_t=W_{\Omega} \mathbf{h}_t$, generating the rotation $\mathbf{R}_t=\exp \left(\left[\boldsymbol{\omega}_t\right]_{\times}\right) \in SO(3)$. Applying this to the normalized pose yields:
\begin{equation}
\mathbf{\bar{M}}_t^{\mathrm{rig}}= \mathbf{R}_t  \mathbf{\bar{M}}_t.
\end{equation}
The residual head produces $\Delta \overline{\mathbf{y}}_t=W_Y \mathbf{h}_t$, giving the refined normalized vector:
\begin{equation}
\widehat{\overline{\mathbf{y}}}_t=\operatorname{vec}\left( \mathbf{\bar{M}}_t^{\mathrm{rig}}\right)+\Delta \overline{\mathbf{y}}_t,
\end{equation}
which is mapped back to the original global frame via $\widehat{\mathbf{M}}_t=s_t \operatorname{unvec}\left(\widehat{\overline{\mathbf{y}}}_t\right)+\mathbf{c}_t \mathbf{1}^{\top} \in \mathbb{R}^{3 \times J}$. This decomposition forces the model to attribute spurious view-induced re-orientations to the rigid head while reserving the residual capacity exclusively for true articulated correction.

Our formulation guarantees analytical translation equivariance via the centroid subtraction and addition process, and encourages full $SE(3)$ equivariance through a label-free geometric loss. To enforce $SO(3)$ equivariance, we supply a geometric signal. Given a random rigid transform $\left( \mathbf{R}_\epsilon, \mathbf{t}_\epsilon\right)$, let $ \mathbf{M}_t^{\prime}=\mathbf{R}_\epsilon \mathbf{M}_t+\mathbf{t}_\epsilon \mathbf{1}^{\top}$ be the transformed input, and $\widehat{\mathbf{M}}_t^{\prime}$ be the rectified output produced by feeding $\left\{\mathbf{M}_t^{\prime}\right\}$ into the module. We define the equivariance loss as:
\begin{equation}
\mathcal{L}_{\text {equiv }}=\frac{1}{T} \sum_{t=1}^T\left\| \widehat{\mathbf{M}}_t^{\prime}-\left( \mathbf{R}_\epsilon \widehat{\mathbf{M}}_t+\mathbf{t}_\epsilon \mathbf{1}^{\top}\right)\right\|_2^2.
\end{equation}

To enforce temporal smoothness without inappropriately penalizing the global movement trajectory of the subject, the kinematic losses strictly operate on the normalized pose coordinates:
\begin{equation}
\mathcal{L}_{\mathrm{vel}}=\frac{1}{T-1} \sum_{t=2}^T\left\| \widehat{\overline{\mathbf{y}}}_t - \widehat{\overline{\mathbf{y}}}_{t-1}\right\|_2^2
\end{equation}
\begin{equation}
\mathcal{L}_{\text {jerk }}=\frac{1}{T-3} \sum_{t=4}^T\left\|\widehat{\overline{\mathbf{y}}}_t - 3 \widehat{\overline{\mathbf{y}}}_{t-1} + 3 \widehat{\overline{\mathbf{y}}}_{t-2} - \widehat{\overline{\mathbf{y}}}_{t-3}\right\|_2^2
\end{equation}

The overall objective:
\begin{equation}
\mathcal{L}_{\mathrm{TER}}=\lambda_{\mathrm{vel}} \mathcal{L}_{\mathrm{vel}}+\lambda_{\mathrm{jerk}} \mathcal{L}_{\mathrm{jerk}}+\lambda_{\mathrm{equiv}} \mathcal{L}_{\mathrm{equiv}},
\end{equation}
where $\lambda_{\mathrm{vel}}$, $\lambda_{\mathrm{jerk}}$, and $\lambda_{\mathrm{equiv}}$ represent balance weights. This objective is fully differentiable and depends only on the inputs $\left\{ \mathbf{M}_t\right\}$. The innovation-gated update mechanism preserves long-term kinematic consistency while suppressing transient observational noise. Simultaneously, the equivariant framework prevents extrinsic camera-induced or subject-induced rigid motions from confounding the intrinsic articulated kinematics.

\section{Experiment}
We systematically conduct experiments on two pubilc datasets Human3.6M \cite{ionescu2013human3} and CMU Panoptic \cite{Joo_2017_TPAMI} to evaluate the performance of the proposed method. The experimental results show the proposed method has achieved more significant performance than existing methods. Considering the clarity of experiments, we also conduct ablation studies on the hyper-parameters and network architecture.

\subsection{Datasets}
\subsubsection{Human3.6M} The Human3.6M dataset is a large commonly used 3D human pose benchmark for both monocular and multi-view, which provided joint keypoint 3D coordinate captured by the MoCap system. The full dataset consists of 11 subjects (S1, S5, S6, S7, S8, S9, S11) which include 15 behaviors such as eating, greeting and sitting. The image frames are collected with 4 synchronized high-resolution digital cameras which are precisely calibrated. Both the digital cameras and the MoCap system are on same clock to ensure perfect synchronization between image frame and joint keypoint 3D coordinate. Following previous works \cite{iskakov2019learnable}, \cite{gordon2022flex}, \cite{jiang2023probabilistic}, we use 5 subjects (S1, S5, S6, S7, S8) for training and 2 subjects (S9, S11) for testing. The 17-joint MPJPE (Mean Per Joint Position Error) is regarded as evaluation metric. We reported the results of the proposed method trained on 4 views.

\subsubsection{CMU Panoptic} The CMU Panoptic contains a large number of RGB images collected by 31 cameras with useful data annotations include camera parameters, 3D poses and 2D poses. We used 17 subjects with COCO format 19-joint annotation for train and evaluation, there is one person in the scene at each point time. Specifically, the dataset is splited by camera views, 4 views for evaluation and up to 27 views for train (some subjects exist data loss). Additionally, we obtain human bounding boxes to extract human regions from the original image, following the approach in \cite{iskakov2019learnable}. The 19-joint MPJPE is regarded as evaluation metric for this dataset.

\subsection{Implementation Details} The method presented in this paper is implemented using the PyTorch framework on 4 $\times$ NVIDIA RTX 4090 GPUs equipped with 24GB memory. The 2D backbone for our experiments was pretrained on the COCO dataset. We used Adam with $4 \times 10^{-4} $ learning rate to optimize the model on this two public datasets. Additionally, we did not use the camera parameters (Intrinsic Parameters and Extrinsic Parameters) of both Human3.6M and CMU Panoptic, which is different from existing methods. 

\begin{table*}[b!]
	\centering
	\renewcommand{\arraystretch}{1}
	\caption{Results of the evaluation on the Human3.6M dataset. The table shows the MPJPE in millimeters for the published state-of-the-art calibrated and uncalibrated methods.
		$\bullet$: calibrated methods. 
		$\diamond$: without extrinsic parameters methods.
		$\circ$:  without intrinsic and extrinsic parameters methods.
		$\star$: Used temporal information.
		\underline{underline}: used 2d keypoints ground-truth.
		The first and second best results of uncalibrated methods are marked in red and blue, respectively. The first best results of calibrated methods in green.
	}
	\resizebox{\textwidth}{!}{
		\begin{tabular}{l|ccccccccccccccc|c}
			\toprule
			Methods & Dir. & Disc.& Eat & Greet & Phone & Photo & Pose & Purch. & Sit & SitD. & Smoke & Wait &  WalkD. & Walk  & WalkT. & Mean \\ \hline 
			\hline
			\multicolumn{17}{c}{Multi-view methods with intrinsic and extrinsic parameters }\\
			\hline
			Tome \textit{et al.} \cite{tome2018rethinking} $\bullet$ &  43.3 &  49.6 &  42.0 &  48.8 &  51.1 &  64.3 &  40.3 &  43.3 &  66.0 &  95.2 & 50.2 & 52.2 & 51.1 & 43.9 & 45.3 & 52.8\\
			Kadkho. \textit{et al.} \cite{kadkhodamohammadi2021generalizable} $\bullet$ & 39.4&46.9&41.0&42.7&53.6&54.8&41.4&50.0&59.9&78.8&49.8&46.2&51.1&40.5&41.0&49.1\\

			Jia \textit{et al.} \cite{jia2023delving} $\bullet$ & - & - & - & - & - & - & - & - &- & - & - & - & - & - & - & 33.0 \\

			Remelli \textit{et al.} \cite{remelli2020lightweight} $\bullet$  & 27.3 &32.1 &25.0 &26.5 &29.3 &35.4 &28.8 &31.6 &36.4 &31.7 &31.2& 29.9 &26.9 &33.7 &30.4 &30.2\\
			Bartol \textit{et al.} \cite{bartol2022generalizable} $\bullet$  
			& 27.5 & 28.4 & 29.3 & 27.5 & 30.1 & 28.1 & 27.9 & 30.8 & 32.9 & 32.5 & 30.8 & 29.4 & 28.5 &  30.5 & 30.1 & 29.1\\
			He \textit{et al.} \cite{he2020epipolar} $\bullet$   &25.7 &27.7 &23.7 &24.8& 26.9 &31.4 &24.9 &26.5 &28.8 &31.7 &28.2 &26.4& 23.6 &28.3 &23.5 &26.9\\
			Qiu \textit{et al.} \cite{qiu2019cross} $\bullet$   & 24.0 &26.7 &23.2 &24.3 &24.8 &22.8& 24.1& 28.6 &32.1 &26.9 &31.0 &25.6 &25.0 &28.0 &24.4 &26.2\\
			Ma \textit{et al.} \cite{ma2021transfusion} $\bullet$ & 24.4& 26.4 &23.4 &21.1 &25.2 &23.2 &24.7& 33.8& 29.8 & 26.4 &26.8& 24.2& 23.2 &26.1 &23.3 &25.8\\
			Iskakov \textit{et al.} \cite{iskakov2019learnable} $\bullet$   &
			19.9 & 20.0 & 18.9  & 18.5 & 20.5 & 19.4  & 18.4 & 22.1  & 22.5  & 28.7 & 21.2 & 20.8  & 19.7  & 22.1  & 20.2  & 20.8\\ 
			Zhang \textit{et al.} \cite{zhang2021adafuse}  $\bullet$ & 
			17.8 & 19.5 & 17.6  & 20.7 & 19.3 &\textcolor{green}{\textbf{16.8}}  &  18.9 & 20.2  & 25.7 &  \textcolor{green}{\textbf{20.1}} & 19.2 &  20.5  & \textcolor{green}{\textbf{17.2}}  & 20.5  &  17.3  & 19.5\\ 
			
			Zhang \textit{et al.} \cite{zhang2024geometry} $\bullet$ & \textcolor{green}{\textbf{ 15.7}}  & \textcolor{green}{\textbf{ 17.4}}  & \textcolor{green}{\textbf{ 17.5}} & \textcolor{green}{\textbf{ 16.0}}  & \textcolor{green}{\textbf{ 18.9}}  & 18.9 & \textcolor{green}{\textbf{15.2}}  & \textcolor{green}{\textbf{16.2}} & \textcolor{green}{\textbf{21.5}} & 22.0 & \textcolor{green}{\textbf{17.8}}  & \textcolor{green}{\textbf{15.6}} & 18.0 & \textcolor{green}{\textbf{14.7}}  & \textcolor{green}{\textbf{16.0}}  & \textcolor{green}{\textbf{17.4}} \\

			\hline
			\multicolumn{17}{c}{Multi-view methods without extrinsic parameters}\\
			\hline
			
			Gordon \textit{et al.} \cite{gordon2022flex} $\diamond \star$
			& \textcolor{cyan}{\textbf{22.0}} & \textcolor{cyan}{\textbf{23.6}} &  24.9 & 26.7  & 30.6
			& 35.7 & 25.1 & 32.9 & 29.5  & 32.5 & 32.6 & 26.5& 34.7 & 26.0 & 27.7 & 30.2\\

			Jiang \textit{et al.} \cite{jiang2023probabilistic} $\diamond$ &24.0&25.4&26.6&30.4&32.1&\textcolor{cyan}{\textbf{20.1}}&\textcolor{cyan}{\textbf{20.5}}&36.5&40.1&\textcolor{cyan}{\textbf{29.5}}&27.4&27.6&\textcolor{cyan}{\textbf{20.8}}&24.1&\textcolor{cyan}{\textbf{22.0}}&27.8\\
			
			Zhang \textit{et al.} \cite{zhang2025efmk} $\diamond$ &23.0& 24.9 &\textcolor{cyan}{\textbf{23.7}} &\textcolor{cyan}{\textbf{22.6}} & \textcolor{cyan}{\textbf{26.6}}& 25.6& 22.1& \textcolor{cyan}{\textbf{25.2}}& \textcolor{cyan}{\textbf{27.4}}& 33.3& \textcolor{cyan}{\textbf{24.7}}&  \textcolor{cyan}{\textbf{22.8}}& 26.7& \textcolor{cyan}{\textbf{22.8}}& 23.1& \textcolor{cyan}{\textbf{25.0}}\\

			\hline
			\multicolumn{17}{c}{Multi-view methods without intrinsic and extrinsic parameters}\\
			\hline
			
			Jia \textit{et al.} \cite{jia2023delving} $\circ$ & - & - & - & - & - & - & - & - &- & - & - & - & - & - & - & 44.8 \\
			
			Huang \textit{et al.} \cite{huang2020deepfuse} $\circ$ &26.8 & 32.0 & 25.6 & 52.1 & 33.3 & 42.3 & 25.8 & 25.9 & 40.5 & 76.6 & 39.1 & 54.5 & 35.9 & 25.1 & 24.2 & 37.5 \\

			Zhang \textit{et al.} \cite{zhang2024deep} $\circ$ &- &- &- &- &- &-& -& -& -& -& -& -& -& -& -& 29.6 \\
			
			Shuai \textit{et al.} \cite{shuai2022adaptive} $\circ$ &24.2&26.4&26.1&25.6&29.4 &29.7&25.1& 25.4& 32.4&37.4&27.1& 25.4&29.5&  23.8 &24.4&27.5\\

			Zhang \textit{et al.} \cite{zhang2025esmformer} $\circ \star$ & 24.0 & 27.5 & 24.8 & 25.0 & 28.8 & 31.3 & 24.7 & 27.0 & 30.6 & 36.2 & 28.6 & 25.3 & 30.0 & 24.7 & 24.0 & 27.4 \\

			Cai \textit{et al.} \cite{cai2024fusionformer} $\circ$ &- &- &- &- &- &- &- &- &- &- &- &- &- &- &- &27.3\\
			
			Zhang \textit{et al.} \cite{zhang2025efmk} $\circ$ &25.1 &26.0 &24.0 &25.0 &26.5 &26.5 &23.5 &27.6 &\textcolor{red}{\textbf{27.7}} &31.0 &25.0 &24.2 &27.7 &24.1 &24.2 &25.8\\
			
			Cai \textit{et al.} \cite{cai2024poseirm} $\circ \star$ & 22.2 & 24.6 & 22.9 & 23.2 &  26.0& 27.0 & \textcolor{blue}{\textbf{22.2}} & \textcolor{blue}{\textbf{23.7}}  & 29.3 & 33.6 & 25.6 & 22.8  & 25.8 & \textcolor{blue}{\textbf{22.5}} & 22.5 & 25.1\\
			
			Ma \textit{et al.} \cite{ma2022ppt} $\circ$ & 21.8  &26.5 & 21.0 & 22.4& 23.7& 23.1 &23.2& 27.9 &30.7 & 24.6 &26.7 &23.3 & 21.2 &25.3 &22.6 & 24.4 \\
			
			\textbf{Our Method} $\circ$ &\textcolor{blue}{\textbf{19.4}}&\textcolor{blue}{\textbf{22.7}}&\textcolor{blue}{\textbf{20.6}}&\textcolor{blue}{\textbf{19.7}}&\textcolor{blue}{\textbf{21.5}}&\textcolor{blue}{\textbf{19.8}}&22.9& 25.2 &32.8&\textcolor{blue}{\textbf{23.0}}&\textcolor{blue}{\textbf{21.6}}&\textcolor{blue}{\textbf{20.9}}&\textcolor{blue}{\textbf{20.5}}& 23.8&\textcolor{blue}{\textbf{21.2}}&\textcolor{blue}{\textbf{22.5}}\\

			\textbf{Our Method} $\circ \star$
			& \textcolor{red}{\textbf{17.5}} & \textcolor{red}{\textbf{19.8}} & \textcolor{red}{\textbf{18.2}} & \textcolor{red}{\textbf{17.6}} & \textcolor{red}{\textbf{18.9}} & \textcolor{red}{\textbf{17.4}} & \textcolor{red}{\textbf{19.5}} & \textcolor{red}{\textbf{22.1}} & \textcolor{blue}{\textbf{28.5}} & \textcolor{red}{\textbf{20.2}} & \textcolor{red}{\textbf{18.7}} & \textcolor{red}{\textbf{18.3}} & \textcolor{red}{\textbf{17.9}} & \textcolor{red}{\textbf{20.5}} & \textcolor{red}{\textbf{18.6}} & \textcolor{red}{\textbf{19.6}} \\
			
			\hline
			
			\multicolumn{17}{c}{Multi-view methods without intrinsic and extrinsic parameters using 2D GT}\\
			\hline

			\underline{Luvizon \textit{et al.}}   \cite{luvizon2022consensus} $\circ$ & 40.0 & 36.0 &44.0& 39.0 &44.0 &42.0 &41.0 &66.0 &70.0 &46.0 &49.0 &43.0 &34.0 &46.0 &34.0 & 45.0 \\ 
			\underline{Gordon \textit{et al.}}  \cite{gordon2022flex}  $\circ \star$ & - & - & - & - & - & - & - & - &- & - & - & - & - & - & - & 22.9 \\
			\underline{Zhang \textit{et al.}}  \cite{zhang2025esmformer} $\circ \star$ & \underline{\textcolor{blue}{\textbf{16.5}}}  & \underline{\textcolor{blue}{\textbf{18.3}}} & \underline{\textcolor{blue}{\textbf{15.1}}} & \underline{\textcolor{blue}{\textbf{17.2}}} & \underline{\textcolor{blue}{\textbf{17.0}}} & \underline{\textcolor{blue}{\textbf{19.4}}} & \underline{\textcolor{blue}{\textbf{16.8}}} & \underline{\textcolor{blue}{\textbf{18.7}}} & \underline{\textcolor{blue}{\textbf{16.4}}} & \underline{\textcolor{blue}{\textbf{22.3}}} & \underline{\textcolor{blue}{\textbf{17.5}}} & \underline{\textcolor{blue}{\textbf{17.3}}} & \underline{\textcolor{blue}{\textbf{20.0}}} & \underline{\textcolor{blue}{\textbf{15.4}}} & \underline{\textcolor{blue}{\textbf{16.8}}} & \underline{\textcolor{blue}{\textbf{17.6}}} \\
			
			\underline{\textbf{Our Method}} $\circ $ & 17.3 & 20.0 & 16.6 & 18.7 & 18.4 & 21.9 & 17.8 & 20.4 & 17.2 & 24.1 & 19.3 & 18.8 & 22.3 & 16.3 & 18.4& 19.2 \\
			
			\underline{\textbf{Our Method}} $\circ \star$ & \underline{\textcolor{red}{\textbf{14.6}}}   & \underline{\textcolor{red}{\textbf{15.8}}} & \underline{\textcolor{red}{\textbf{13.0}}} & \underline{\textcolor{red}{\textbf{15.9}}} & \underline{\textcolor{red}{\textbf{14.3}}} & \underline{\textcolor{red}{\textbf{17.2}}} & \underline{\textcolor{red}{\textbf{14.7}}} & \underline{\textcolor{red}{\textbf{15.4}}} & \underline{\textcolor{red}{\textbf{14.2}}} & \underline{\textcolor{red}{\textbf{18.6}}} & \underline{\textcolor{red}{\textbf{15.2}}} & \underline{\textcolor{red}{\textbf{15.7}}} & \underline{\textcolor{red}{\textbf{17.3}}} & \underline{\textcolor{red}{\textbf{13.2}}} & \underline{\textcolor{red}{\textbf{14.1}}} & \underline{\textcolor{red}{\textbf{15.3}}} \\
			
			\bottomrule

		\end{tabular}
	}
	\label{tab1}
	
\end{table*}

\subsection{Comparison Results} 
\subsubsection{Results on Human3.6M} In Table~\ref{tab1}, we report quantitative results on the Human3.6M dataset, comparing our approach against a broad range of state-of-the-art multi-view methods. Among methods utilizing full calibration, the geometry-aware approach of Zhang \textit{et al.}~\cite{zhang2024geometry} achieves the lowest mean error of 17.4 mm, setting the current benchmark for this setting. This method significantly improves upon earlier baselines, such as AdaFuse~\cite{zhang2021adafuse} and the learnable triangulation of Iskakov \textit{et al.}~\cite{iskakov2019learnable}. These results demonstrate that with accurate camera parameters, advanced geometric reasoning can drive the error well below 20 mm.

For methods relying solely on intrinsic parameters, performance generally bridges the gap between the fully calibrated and uncalibrated regimes. Within this group, EFMK~\cite{zhang2025efmk} attains the best mean error of 25.0 mm, outperforming Flex~\cite{gordon2022flex} and the probabilistic model of Jiang \textit{et al.}~\cite{jiang2023probabilistic}. Remarkably, despite strictly avoiding any camera parameters, our method achieves a lower mean error of 19.6 mm, highlighting the efficacy of our proposed uncalibrated formulation over intrinsics-dependent baselines.

In the most challenging scenario characterized by the absence of both intrinsic and extrinsic parameters, our method delivers the best overall performance with a mean MPJPE of 19.6 mm. It outperforms the previous leading method PPT~\cite{ma2022ppt} which reports 24.4 mm by a margin of 4.8 mm. This corresponds to a relative improvement of approximately 19\%. Other recent methods like PoseIRM~\cite{cai2024poseirm} and EFMK~\cite{zhang2025efmk} trail with mean errors of 25.1 mm and 25.8 mm respectively. Our approach achieves the lowest error in 14 out of 15 action categories. These gains are particularly pronounced for actions involving large pose variations or self-occlusion such as \textit{Phone, Photo, Smoke,} and \textit{WalkDog}. For instance, the error on \textit{Smoke} drops from the 26.7 mm reported by PPT to 18.7 mm. The only notable limitation appears in the \textit{Sit} category where our method remains competitive but has not yet surpassed EFMK~\cite{zhang2025efmk}.

When ground-truth 2D keypoints are provided, the performance of all methods improves significantly. Under this setting, ESMFormer~\cite{zhang2025esmformer} reaches a mean error of 17.6 mm and serves as a strong baseline among uncalibrated methods that exploit temporal information. Our single-frame variant attains 19.2 mm and is therefore slightly weaker than ESMFormer when temporal cues are not used. Once temporal module TER is enabled, our method achieves a mean MPJPE of 15.3 mm and establishes a new state-of-the- art. This result surpasses ESMFormer by 2.3 mm which corresponds to a relative gain of about 13\%. A direct comparison between our own two variants also highlights the effect of temporal information. The mean MPJPE decreases from 19.2 mm to 15.3 mm to yield an additional reduction of roughly 20\%. Overall, even without any camera calibration, the proposed method approaches the accuracy of recent calibrated approaches and highlights its potential for practical and easy-to-deploy multi-camera setups.

\begin{table}[t!]
	\caption{Results of the evaluation on the CMU Panoptic dataset. $\bullet$: calibrated methods. 
	$\diamond$: without extrinsic parameters methods.
	$\circ$:  without intrinsic and extrinsic parameters methods.
	$\star$: Used temporal information. (Using 4 cameras)}
	\label{tab2}
	\centering
	\renewcommand{\arraystretch}{1}
	\footnotesize
	\tabcolsep 7pt
	\begin{tabular}{c|c}
		\hline 
		Methods & Mean (MPJPE, mm) \\ \toprule
		\multicolumn{2}{c}{Multi-view methods with intrinsic and extrinsic parameters }\\
		\hline
		Iskakov \textit{et al.} Algebraic \cite{iskakov2019learnable} $\bullet$  & 21.3 \\
		Iskakov \textit{et al.} Volumetric \cite{iskakov2019learnable} $\bullet$  & 13.7 \\ 
		Ye \textit{et al.} \cite{ye2022faster} $\bullet$  & 17.9 \\
		Liao \textit{et al.} \cite{liao2024multiple} $\bullet$  & 15.6 \\
		Chharia \textit{et al.} \cite{chharia2025mv} $\bullet$  & 15.3 \\
		Zhang \textit{et al.} \cite{zhang2024geometry} $\bullet$   & \textcolor{green}{\textbf{11.2}} \\
		\hline
		\multicolumn{2}{c}{Multi-view methods without extrinsic parameters}\\
		\hline
		Gordon \textit{et al.}   \cite{gordon2022flex} $\diamond$ & 28.9 \\
		Bartol \textit{et al.}   \cite{bartol2022generalizable} $\diamond$ & 25.4 \\
		Jiang \textit{et al.}  \cite{jiang2023probabilistic} $\diamond$ & \textcolor{cyan}{\textbf{24.2}} \\
		\hline
		\multicolumn{2}{c}{Multi-view methods without intrinsic and extrinsic parameters}\\
		\hline
		Our Method $\circ$ & \textcolor{blue}{\textbf{12.8}} \\
		Our Method $\circ \star$ & \textcolor{red}{\textbf{10.6}} \\ 		\bottomrule

	\end{tabular}
\end{table}

\subsubsection{Results on CMU Panoptic} As shown in Table~\ref{tab2}, among fully calibrated approaches, the geometry-aware method of Zhang \textit{et al.} \cite{zhang2024geometry} achieves the best performance with a mean error of 11.2 mm. This result is substantially lower than earlier calibrated baselines such as the volumetric variant of learnable triangulation by Iskakov \textit{et al.} \cite{iskakov2019learnable} at 13.7 mm. It also outperforms more recent approaches by Liao \textit{et al.} \cite{liao2024multiple} and Chharia \textit{et al.} \cite{chharia2025mv} which report errors of 15.6 mm and 15.3 mm respectively. These results indicate that strong geometric modeling can reduce the error to nearly 11 mm on the CMU Panoptic dataset when full calibration is available.

Performance degrades noticeably when extrinsic calibration is removed even if intrinsic parameters are preserved. Within this category, the method of Jiang \textit{et al.} \cite{jiang2023probabilistic} records an error of 24.2 mm while the approaches of Gordon \textit{et al.} \cite{gordon2022flex} and Bartol \textit{et al.} \cite{bartol2022generalizable} obtain 28.9 mm and 25.4 mm respectively. This significant performance gap suggests that partial calibration is insufficient to fully exploit multi-view consistency on this dataset. Our method addresses the most challenging scenario characterized by the complete absence of intrinsic and extrinsic parameters. Even in its single-frame configuration, it attains a mean error of 12.8 mm. This result is clearly superior to all methods relying solely on intrinsics and is already comparable to top-tier calibrated approaches. Upon incorporating temporal information, our method achieves 10.6 mm to set a new state-of-the-art on the CMU Panoptic dataset with 4 cameras. Notably, this temporal variant surpasses the strongest calibrated baseline by Zhang \textit{et al.} \cite{zhang2024geometry} and demonstrates that a fully uncalibrated formulation can match or even exceed the accuracy of recent calibrated systems while offering a significantly simpler deployment pipeline.

\begin{figure}[b!]
	\begin{center}
		\includegraphics[width = 1\linewidth]{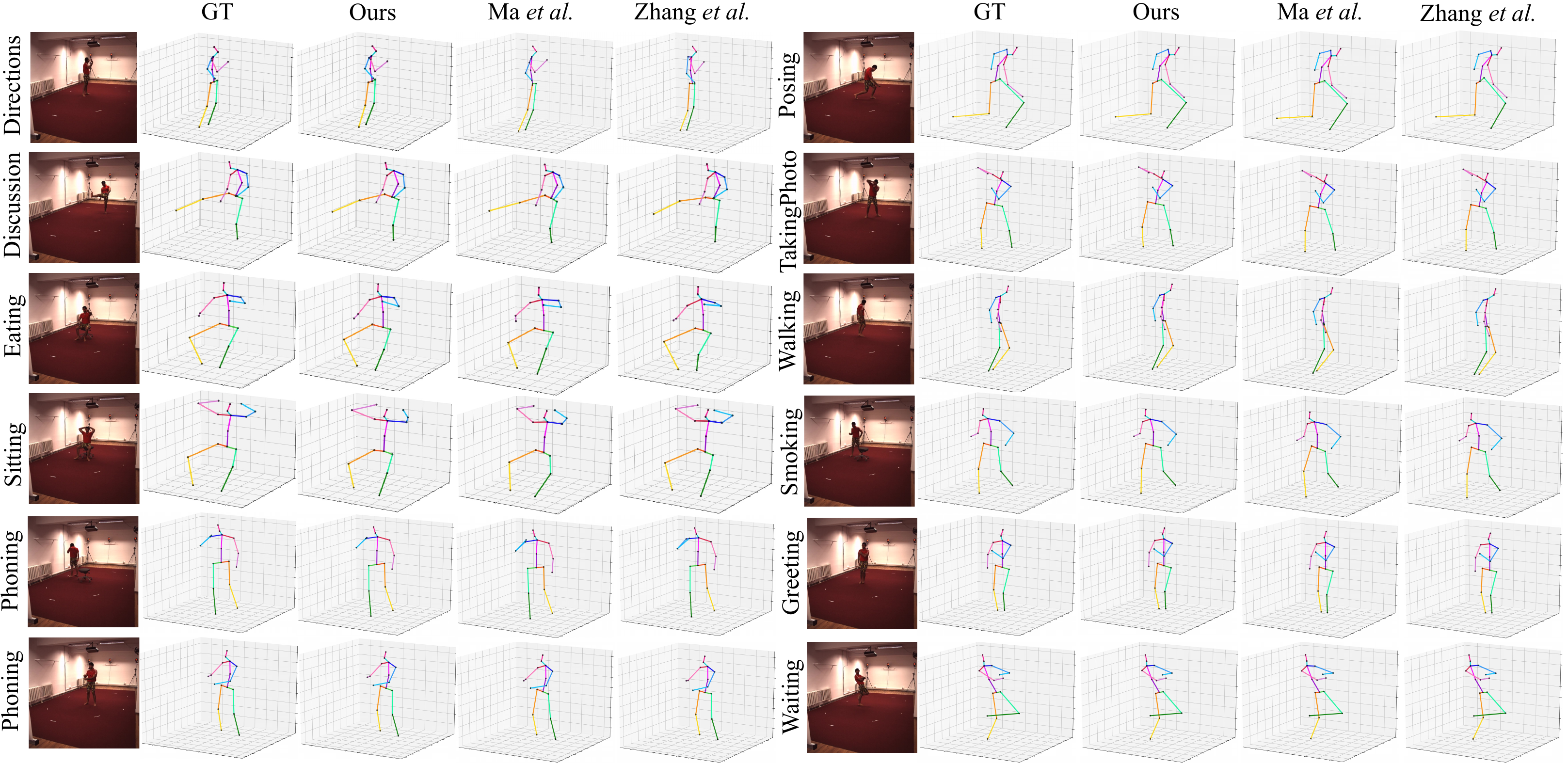}
	\end{center}
	\caption{Results visualization of our method on Human3.6M dataset. We compared with Ma \textit{et al.} \cite{ma2022ppt} and Zhang \textit{et al.} \cite{zhang2025efmk}.}
	\label{human3.6}
\end{figure}

\subsubsection{Result Visualization on Human3.6M}
To evaluate the kinetic versatility of our framework, we present a comprehensive gallery of 3D pose reconstructions on the Human3.6M benchmark. Figure \ref{human3.6} provides visual evidence across a remarkably diverse array of everyday activities. These sequences span the entire kinetic spectrum, ranging from highly dynamic motions like walking and greeting to relatively static postures such as sitting and waiting. Across this extensive variety of actions, the proposed method yields pose estimates that align exceptionally well with the ground-truth.

A closer examination against contemporary state-of-the-art models, specifically Ma \textit{et al.} \cite{ma2022ppt} and Zhang \textit{et al.},  \cite{zhang2025efmk} reveals the distinct advantages of our approach in maintaining kinematic fidelity. In scenarios characterized by severe self-occlusion, such as the phoning and taking-photo sequences, the subject's hands and arms frequently obstruct the torso from the camera's line of sight. Under such conditions, competing methods often struggle to resolve depth ambiguities, leading to misaligned limbs or structurally incorrect joint placements. Conversely, our model successfully disambiguates these complex overlapping regions, preserving the precise articulation of the limbs while maintaining overall structural integrity.

The superiority of these reconstructions can be directly attributed to the well-conditioned optimization landscape formulated by our GC module and the robust noise rejection mechanism of the TER. While baseline models exhibit noticeable positional deviations in distal joints, such as wrists and ankles, due to the accumulation of uncalibrated triangulation errors, our framework effectively filters out these view-induced perturbations. The resulting skeletons consistently demonstrate accurate global orientations and physically plausible joint angles, confirming the method's reliability in handling the extensive behavioral variability inherent in natural human movement.

\begin{figure}[t!]
	\begin{center}
		\includegraphics[width = 1\linewidth]{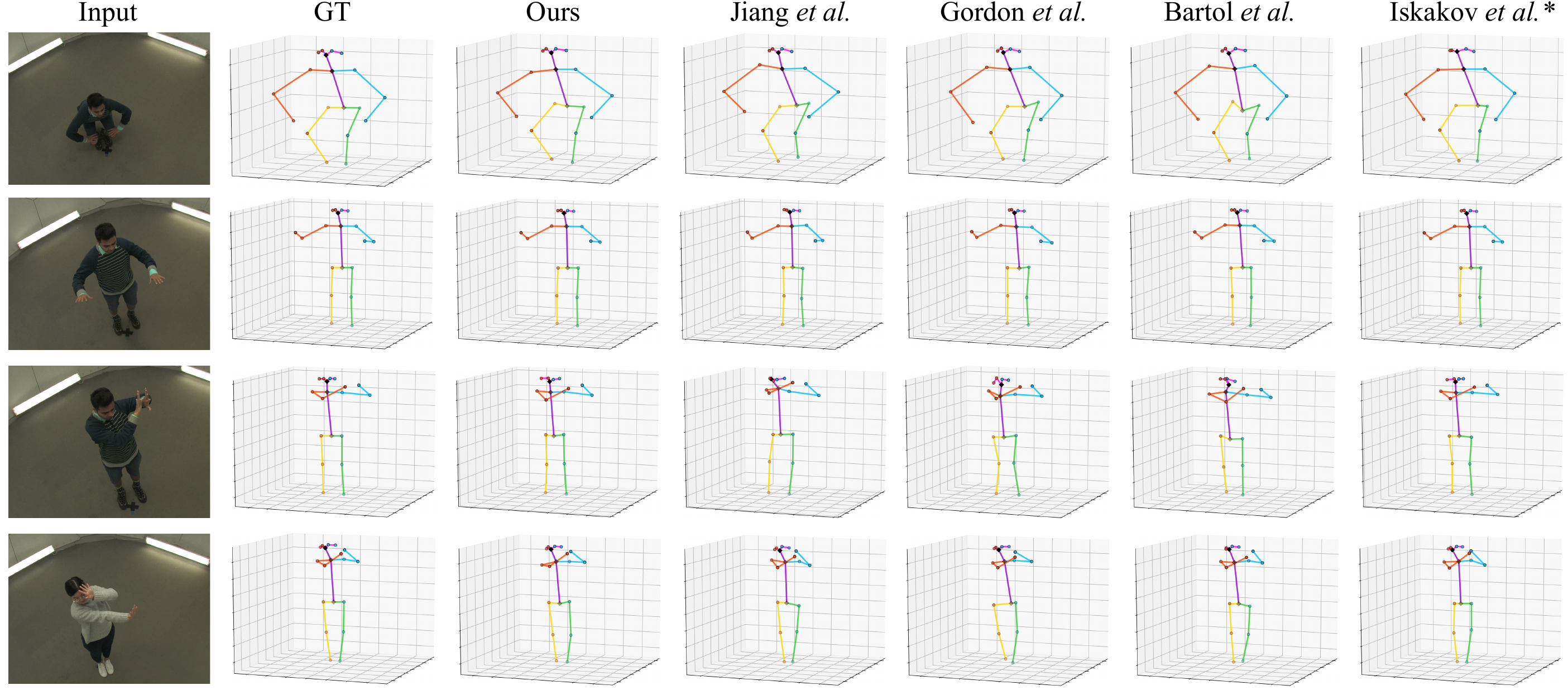}
	\end{center}
	\caption{Results visualization of our method on CMU Panoptic dataset. We compared with Jiang \textit{et al.} \cite{jiang2023probabilistic},  Gordon \textit{et al.} \cite{gordon2022flex}, Bartol \textit{et al.} \cite{bartol2022generalizable} and Iskakov \textit{et al.} \cite{iskakov2019learnable}. *: Calibrated methods.}
	\label{cmu}
\end{figure}
\subsubsection{Result Visualization on CMU Panoptic}
To further assess the generalization capability and robustness of our framework under challenging camera viewpoints, we visualize the 3D pose reconstruction results on the CMU Panoptic dataset. Unlike standard frontal-view sequences, this dataset frequently features extreme high-angle and top-down perspectives. These viewpoints introduce severe perspective distortion and depth foreshortening, making it exceedingly difficult to infer accurate 3D geometry from 2D observations. 

As illustrated in Figure \ref{cmu}, our proposed method maintains exceptional spatial congruence with the ground-truth, successfully recovering the canonical 3D structure despite the skewed camera pitch. In contrast, state-of-the-art baselines, including Jiang \textit{et al.} \cite{jiang2023probabilistic}, Gordon \textit{et al.} \cite{gordon2022flex}, Bartol \textit{et al.} \cite{bartol2022generalizable}, and Iskakov \textit{et al.} \cite{iskakov2019learnable}, struggle significantly with these extreme views. For instance, in the top-down sequences where the subject is reaching outward (rows 1 and 2), the heavy foreshortening causes competing methods to suffer from severe depth ambiguity. Consequently, this leads to twisted limb orientations, out-of-plane joint predictions, and anatomically implausible bone lengths, which is particularly evident in the severely distorted arm estimations generated by Jiang \textit{et al.} \cite{jiang2023probabilistic} and Gordon \textit{et al.} \cite{gordon2022flex}. Furthermore, during complex gestures where the hands interact closely with the body or face, such as those depicted in rows 3 and 4, the baseline methods frequently exhibit joint ``collapsing," thereby merging the depth planes of the arms and the torso. Our approach completely avoids these anatomical violations. Driven by the robust geometric regularization of the GC and the adaptive focal cues captured by our dual-token design, the model accurately resolves the projective ambiguities induced by extreme viewpoints. Ultimately, it preserves strict bone-length constancy and maintains the correct spatial separation between the limbs and the torso, proving its superior adaptability in uncalibrated scenarios.

\subsection{Camera Extrinsic Parameters Estimation}

\begin{figure}[t!]
	\begin{center}
		\includegraphics[width = 1\linewidth]{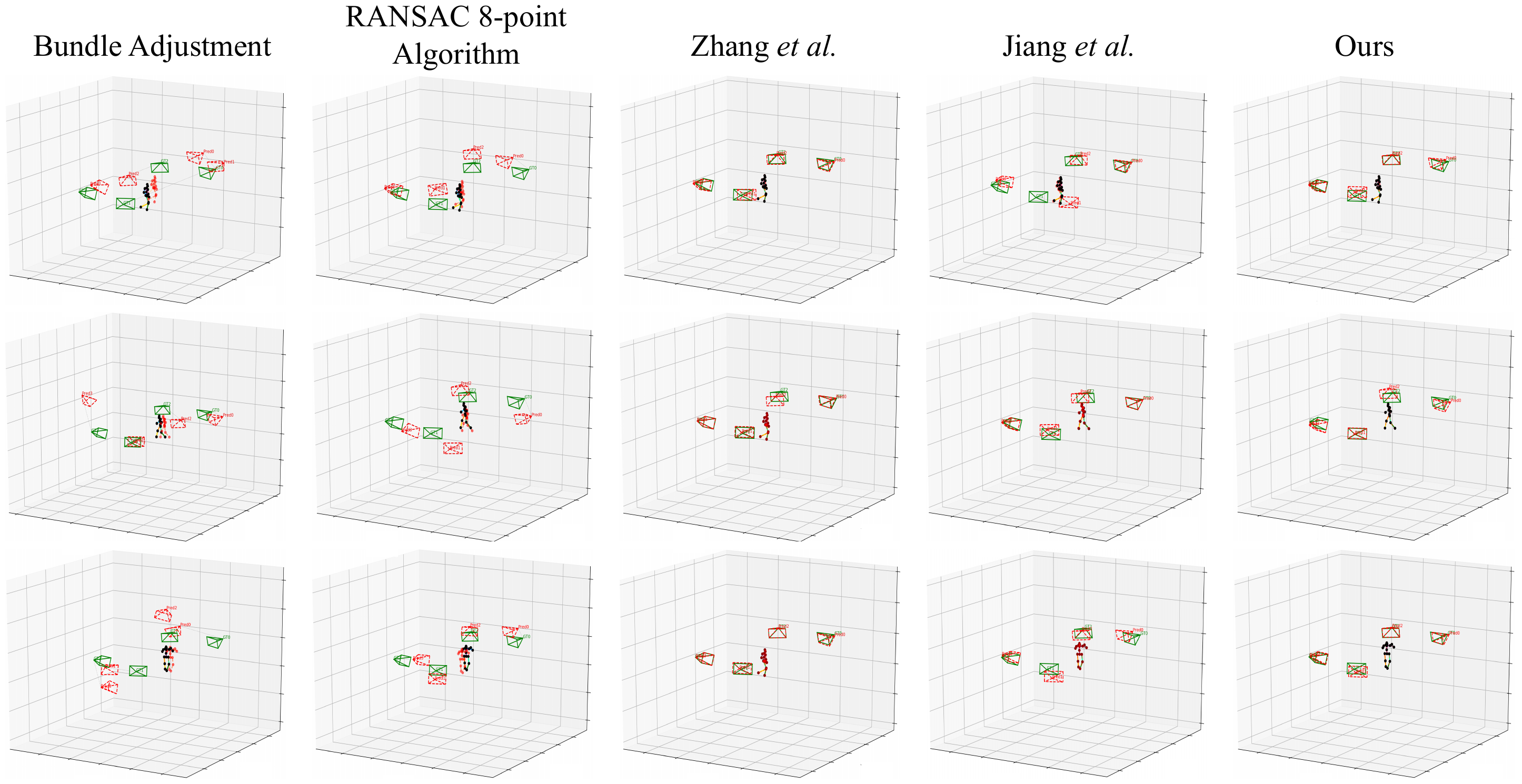}
	\end{center}
	\caption{Comparative 3D human and camera pose estimation on Human3.6M. The Pyramids denote camera pose parameters, the green part represents the ground-truth, and the red part represents the prediction results of different techniques. Baseline methods (Bundle Adjustment, RANSAC 8-point) exhibit significant deviations from ground- truth, while our method achieves superior alignment accuracy.}
	\label{camera}
\end{figure}

In uncalibrated multi-view human pose estimation, recovering the camera extrinsic parameters is as critical as inferring the 3D human joint coordinates. To demonstrate the robustness of our framework in spatial calibration, we conducted a qualitative evaluation of the estimated camera poses on the Human3.6M dataset. Figure \ref{camera} provides a visual comparison of the reconstructed camera frustums against the ground-truth, mapping the predictions from classical geometric algorithms and contemporary deep learning approaches.

Traditional structure-from-motion techniques, specifically the RANSAC 8-point Algorithm and Bundle Adjustment, heavily rely on dense, static, and highly accurate keypoint correspondences. When applied to sparse and noisy 2D human joint detections, these classical methods suffer from severe geometric instability. The non-rigid nature of human articulation violates the static scene assumption, causing the optimization to fall into degenerate configurations. Consequently, this leads to catastrophic failures in extrinsic estimation, which is visually evident in the severe divergence and scattered placement of the predicted camera frustums compared to the ground- truth. 

While recent learning-based methods, including Zhang \textit{et al.} \cite{zhang2025efmk} and Jiang \textit{et al.} \cite{jiang2023probabilistic}, mitigate these classical failures by leveraging data-driven priors, they still exhibit noticeable alignment drift. In these baseline results, the estimated camera orientations frequently show angular deviations, and the translations fail to preserve the correct metric scale relative to the central subject. This indicates that these networks struggle to fully disentangle the global camera extrinsics from the localized human poses. In contrast, our proposed method achieves near-perfect spatial alignment with the ground-truth camera configurations. The predicted camera frustums strictly overlap with the target orientations and positions across all views. This superior calibration accuracy is fundamentally driven by the GC, which enforces strict algebraic multi-view constraints, effectively guiding the optimization process toward the true geometric manifold and bypassing degenerate configurations. Ultimately, this ensures that the camera extrinsic parameters are accurately isolated and recovered from the raw 2D observations, proving the framework's exceptional reliability in fully uncalibrated scenarios.

\subsection{Ablation Studies}

\begin{table}[t!]
	\centering
	\renewcommand{\arraystretch}{1.05}
	\footnotesize
	\setlength{\tabcolsep}{5pt}
	\caption{Ablation study of GC components. We analyze the impact of separate geometric losses and their combinations. Note: This ablation experiment is conducted without the TER module.}
	\label{tab:ablation_GC_grouped}
	\begin{tabular}{clccccc}
		\toprule
		& & \multicolumn{3}{c}{Loss Components} & \multicolumn{2}{c}{MPJPE $\downarrow$} \\
		\cmidrule(lr){3-5} \cmidrule(lr){6-7}
		& Setting & $\mathcal{L}_{\mathrm{GB-2}}$ & $\mathcal{L}_{\mathrm{GB-3}}$ & $\mathcal{L}_{\mathrm{GB-4}}$ & Human3.6M & CMU \\
		\midrule
		\multicolumn{7}{l}{\textit{Baselines}} \\
		(a) & No-geo (Pose only)           &            &            &            & 40.2 & 38.5 \\
		(b) & Sampson distance \cite{hartley2003multiple}             &            &            &            & 25.8 & 14.9 \\
		\midrule
		\multicolumn{7}{l}{\textit{Individual Components}} \\
		(c) & w/ GB-2                      & \checkmark &            &            & 25.3 & 14.7 \\
		(d) & w/ GB-3                      &            & \checkmark &            & 25.9 & 15.1 \\
		(e) & w/ GB-4                      &            &            & \checkmark & 26.2 & 15.4 \\
		\midrule
		\multicolumn{7}{l}{\textit{Combinations}} \\
		(f) & GB-2 + GB-3                  & \checkmark & \checkmark &            & 24.0 & 13.7 \\
		(g) & GB-2 + GB-4                  & \checkmark &            & \checkmark & 24.1 & 13.8 \\
		(h) & GB-3 + GB-4                  &            & \checkmark & \checkmark & 24.5 & 14.1 \\
		\midrule
		(i) & \textbf{Full GC (Ours)}      & \checkmark & \checkmark & \checkmark & \textbf{22.5} & \textbf{12.8} \\
		\bottomrule
	\end{tabular}
\end{table}
\subsubsection{Analysis of GC} 
To thoroughly evaluate the efficacy of the GC, we conduct an ablation study on the Human3.6M and CMU Panoptic datasets while isolating the multilinear geometric constraints from the TER module. The absolute baseline model lacking explicit geometric supervision suffers severe performance degradation with an MPJPE of 40.2 mm on Human3.6M, which empirically validates the necessity of geometric constraints. 

When introducing pairwise supervision, our purely algebraic bilinear formulation ($\mathcal{L}_{\mathrm{GB-2}}$) suppresses the error to 25.3 mm, outperforming the classical first-order Sampson distance ($25.8$ mm). This demonstrates that evaluating universal Gr\"{o}bner basis residuals provides a more stable optimization landscape than traditional local approximations. Interestingly, applying the high-order constraints individually ($\mathcal{L}_{\mathrm{GB-3}}$ or $\mathcal{L}_{\mathrm{GB-4}}$) yields slightly higher errors than the baseline $\mathcal{L}_{\mathrm{GB-2}}$. This performance degradation is theoretically expected, as isolated high-degree polynomial residuals generate severely non-linear gradients that amplify early-stage optimization noise in the absence of lower-order regularization. However, these high-order generators excel at resolving complex projective ambiguities. Ultimately, the true potential of the algebraic regularization is unlocked by aggregating the bilinear, trilinear, and quadrilinear generators into the Full GC model. The foundational bilinear terms provide stable numerical anchors, while the high-order terms enforce global geometric consensus, achieving the best overall performance of $22.5$ mm and $12.8$ mm on the respective datasets.

\begin{table}[b!]
	\centering
	\renewcommand{\arraystretch}{1}
	\footnotesize
	\tabcolsep 37pt
	\caption{Ablation of TTR.
		U: Uncertainty bias. $K$-Mode: Intrinsic configuration ($\text{S}{+}\Delta$: Shared + per-pair update, $\text{S}$: Shared only, $\text{Fix}$: Dataset mean, $\text{Free}$: Independent per-pair). Note: This ablation experiment conducted without TER.
	}
	\label{tab:ablation_transformer_single_col}
	
	\setlength{\tabcolsep}{2.5pt}
	\begin{tabular}{l c c c c c c}
		\toprule
		& \multicolumn{3}{c}{Components} & \multicolumn{1}{c}{Setting} & \multicolumn{2}{c}{MPJPE $\downarrow$} \\
		\cmidrule(lr){2-4} \cmidrule(lr){5-5} \cmidrule(lr){6-7}
		Model Variant & [POSE] & [K] & U & $K$-Mode & Human3.6M & CMU \\
		\midrule
		\textit{Baselines} & & & & & & \\
		(a) Direct-$F$ reg. & & & \checkmark & -- & 29.5 & 17.2 \\
		(b) Direct-3D & & &  & -- & 38.2 & 21.5 \\
		
		\midrule
		\textit{Token Ablation} & & & & & & \\
		(c) No global tokens  & & & \checkmark & Free & 26.1 & 15.4 \\
		(d) Only [POSE] token & \checkmark & & \checkmark & Fix & 25.4 & 14.9 \\
		\midrule
		\textit{Mechanism Ablation} & & & & & & \\
		(e) No Unc. bias  & \checkmark & \checkmark & & $\text{S}{+}\Delta$ & 23.9 & 13.5 \\
		(f) Global-$K$ only & \checkmark & \checkmark & \checkmark & S & 23.2 & 13.1 \\
		(g) Per-pair $K$  & \checkmark & \checkmark & \checkmark & Free & 24.8 & 14.3 \\
		\midrule
		(h) \textbf{Full (Ours)}  & \checkmark & \checkmark & \checkmark & $\text{S}{+}\Delta$ & \textbf{22.5} & \textbf{12.8} \\
		\bottomrule
	\end{tabular}
\end{table}

\subsubsection{Analysis of TTR}
We report the ablation study of our Transformer-based regressor in Table \ref{tab:ablation_transformer_single_col}. First, the baselines underscore the necessity of our explicit geometric modeling. The Direct-3D baseline in row (b) employs a feed-forward network to map 2D inputs directly to 3D coordinates and yields the highest error of 38.2 mm. This confirms that without the inductive bias of triangulation, the model fails to generalize in uncalibrated settings. Similarly, Direct-$F$ regression in row (a) directly predicts the 9-dimensional fundamental matrix without decomposing it into camera parameters and lags significantly behind our full model with an error of 29.5 mm versus 22.5 mm. This suggests that explicitly disentangling intrinsics and extrinsics provides a more stable optimization landscape than learning the epipolar geometry implicitly. Regarding the token design, removing global tokens in row (c) or fixing the intrinsic token in row (d) leads to suboptimal performance, verifying the importance of learnable query tokens for aggregating camera-specific information. A key finding lies in the intrinsic configuration or $K$-Mode. The Free strategy in row (g) predicts independent intrinsics for each pair and performs worse with 24.8 mm error than the Shared strategy in row (f) which achieves 23.2 mm. This indicates that estimating unconstrained parameters for every pair introduces instability and over-parameterization. Our proposed $\text{S}{+}\Delta$ strategy shown in row (h) effectively bridges this gap by regularizing the prediction with a shared base while allowing instance-specific refinements, achieving the best trade-off with the lowest error of 22.5 mm. Finally, the uncertainty bias seen by comparing row (e) against row (h) further improves accuracy by suppressing noisy visual evidence.

\subsubsection{Analysis of TER} 
Table \ref{ab:TER} presents the ablation study for the proposed TER. The frame-wise baseline in row (a) establishes the initial performance at $22.5$ mm on Human3.6M and $12.8$ mm on CMU. While incorporating temporal context generally improves estimation stability, standard recurrent architectures such as the plain GRU in row (b) yield only marginal gains as they struggle with camera-induced rigid motions and transient noise in uncalibrated settings. We further validate the contribution of each specialized design via targeted ablations. Specifically, removing the rigid-non-rigid decomposition head in row (e) leads to the most severe performance degradation, which highlights the critical necessity of disentangling view-induced global re-orientation from articulated deformation. 

Similarly, the ablation of dynamic input features in row (c), innovation gating in row (d), and the $SE(3)$ equivariance loss in row (f) all incur distinct performance drops, thereby verifying their respective roles in modeling short-term motion dynamics, suppressing outlier frames, and enforcing geometrically consistent corrections. Although the kinematic smoothness losses in row (g) show only a minor impact on absolute pose accuracy, they primarily optimize temporal stability rather than per-frame position precision. Finally, our Full TER in row (h) integrates all proposed components to achieve a significant performance gain, reducing the error to $19.6$ mm on Human3.6M and $10.6$ mm on CMU. Notably, the consistent performance across both datasets demonstrates that our geometrically grounded temporal modeling effectively compensates for the lack of static calibration by leveraging dynamic motion cues.

\begin{table}[hp]
	\centering
	\renewcommand{\arraystretch}{1}
	\footnotesize
	\caption{Ablation Study of the Temporal Equivariant Rectifier. We compare our full model against static baselines and ablated variants.}
	\setlength{\tabcolsep}{4pt}
	\begin{tabular}{l c c c c c c c}
		\toprule
		& \multicolumn{5}{c}{TER Components} & \multicolumn{2}{c}{MPJPE $\downarrow$} \\
		\cmidrule(lr){2-6} \cmidrule(lr){7-8}
		Variant & Dynamic Input & Innov. Gate & Rigid-Non-Rigid Decomp. & $\mathcal{L}_{\text{equiv}}$ & $\mathcal{L}_{\text{vel+jerk}}$ & Human3.6M & CMU \\
		\midrule
		\textit{Baselines} & & & & & & & \\
		(a) Baseline (w/o TER) & & & & & & 22.5 & 12.8 \\
		(b) Plain GRU & \checkmark & & & & & 21.5 & 12.0 \\
		\midrule
		\textit{Ablation Study} & & & & & & & \\
		(c) w/o Dynamic Input & & \checkmark & \checkmark & \checkmark & \checkmark & 20.8 & 11.4 \\
		(d) w/o Innov. Gate & \checkmark & & \checkmark & \checkmark & \checkmark & 20.5 & 11.2 \\
		(e) w/o Rigid-Non-Rigid Decomp. & \checkmark & \checkmark & & \checkmark & \checkmark & 21.0 & 11.6 \\
		(f) w/o $\mathcal{L}_{\text{equiv}}$ & \checkmark & \checkmark & \checkmark & & \checkmark & 20.3 & 11.0 \\
		(g) w/o $\mathcal{L}_{\text{vel+jerk}}$ & \checkmark & \checkmark & \checkmark & \checkmark & & 20.0 & 10.9 \\ \hline
		(h) \textbf{Full TER (Ours)} & \checkmark & \checkmark & \checkmark & \checkmark & \checkmark & \textbf{19.6} & \textbf{10.6} \\
		\bottomrule
	\end{tabular}
	\label{ab:TER}
\end{table}

\section{Conclusion}
In this paper, we presented an unconstrained framework for uncalibrated multi-view 3D human pose estimation that successfully eliminates the traditional dependency on precise camera calibration. By synergizing deep representation learning with algebraic geometry, our approach recovers accurate 3D human articulation and spatial camera configurations solely from 2D observations. This is achieved through three core innovations. The Triangulation with Transformer Regressor reformulates multiview fusion into a robust data-driven process. Concurrently, the Gr\"{o}bner basis Corrector translates the algebraic priors into differentiable polynomial constraints, which dynamically projects the uncalibrated estimations onto a geometric space, while the Temporal Equivariant Rectifier disentangles view-induced rigid transformations from articulated dynamics to resolve scale ambiguity and instability.

Extensive evaluations on the Human3.6M and CMU Panoptic datasets demonstrate that our method establishes a new state-of-the-art performance benchmark. The proposed framework consistently yields anatomically plausible and geometrically consistent 3D skeletons under severe perspective distortions and self-occlusions. Ultimately, this work significantly narrows the performance gap between calibration-free algorithms and fully calibrated systems, paving the way for more flexible human-centric motion capture in wild environments. Future research will focus on extending this unified geometric framework to multi-person interactive scenarios and exploring lightweight adaptations for real-time deployment.

\Acknowledgements{This work was supported by the Major Program of the National Natural Science Foundation of China under Grant 62495064, the National Natural Science Foundation of China under Grant No. 62572104, the Sichuan Science and Technology Program (Grant Nos. 2025JDDQ0008, 2024NSFJQ0035), and the Talents by Sichuan provincial Party Committee Organization Department.}



 \bibliographystyle{scis}
 \bibliography{main}

%
%
%



\end{document}